\documentclass{article}

    \PassOptionsToPackage{numbers, compress}{natbib}

 \usepackage[main, final]{neurips_2026}

\usepackage[utf8]{inputenc} 
\usepackage[T1]{fontenc}    
\usepackage{hyperref}       
\usepackage{url}            
\usepackage{booktabs}       
\usepackage{amsfonts}       
\usepackage{xcolor}         
\usepackage{graphicx} 
\usepackage{amsmath}
\usepackage{multirow}
\usepackage{makecell}
\usepackage{caption}
\usepackage{colortbl}

\title{LAFP: Preserving Latent Action Structure in Latent Policy Learning via Flow Matching}

\newcommand{\authorlinesep}{1.5pt}

\author{%
  Jiexi Lyu\thanks{Equal contribution.} \\[\authorlinesep]
  Fudan University \\[\authorlinesep]
  {\scriptsize\texttt{jxlv23@m.fudan.edu.cn}} \\
  \And
  Xizhou Bu\footnotemark[1] \\[\authorlinesep]
  Fudan University \\[\authorlinesep]
  {\scriptsize\texttt{xzbu24@m.fudan.edu.cn}} \\
  \And
  Qingqiu Huang \\[\authorlinesep]
  Morphi Robot \\[\authorlinesep]
  {\scriptsize\texttt{draco.huang@morphi.com}} \\
  \And
  Chufeng Tang \\[\authorlinesep]
  Morphi Robot \\[\authorlinesep]
  {\scriptsize\texttt{felix.tang@morphi.com}} \\
  \AND
  Xiaoshuai Hao \\[\authorlinesep]
  Xiaomi EV \\[\authorlinesep]
  {\scriptsize\texttt{haoxiaoshuai714@163.com}} \\
  \And
  Hongbo Wang \\[\authorlinesep]
  Fudan University \\[\authorlinesep]
  {\scriptsize\texttt{Wanghongbo@fudan.edu.cn}} \\
  \And
  Wei Li\thanks{Corresponding author.} \\[\authorlinesep]
  Fudan University \\[\authorlinesep]
  {\scriptsize\texttt{fd\_liwei@fudan.edu.cn}}
}

\begin{document}

\maketitle

\begin{abstract}
  Learning high quality latent actions from large-scale unlabeled videos, coupled with limited real-world interaction data for training an action decoder, has emerged as a promising paradigm for scalable latent policy learning. However, existing approaches typically rely on behavior cloning, which tends to collapse inherently multimodal action distributions into unimodal ones, thereby degrading the pretrained latent action structure. While flow matching provides a potential alternative, directly applying it leads to a misalignment between latent actions and physical actions during action decoder training, due to the stochastic nature of the learned policy. To address these, we propose Latent Action Flow Policy (LAFP), which leverages flow matching for latent policy learning and introduces an inference-time interpolation mechanism to mitigate stochasticity-induced misalignment. Experimental results demonstrate that LAFP consistently outperforms prior methods on most downstream imitation learning tasks, achieving an 8.2\% improvement in average success rate while incurring less than 1× additional inference overhead.
\end{abstract}

\section{Introduction}

In recent years, the severe scarcity of high quality action data has emerged as a fundamental bottleneck in advancing embodied intelligence. This has motivated a promising paradigm that learns latent actions from large-scale unlabeled video data, which can be naturally formulated as a three-stage pipeline \cite{mccarthy2025towards, nie2026lary, yu2026latent}. In the pretraining stage, approaches such as the Latent Action Policies (LAPO) \cite{schmidt2023learning} adopt an autoencoding framework, where an inverse dynamics model (IDM) and a forward dynamics model (FDM) are jointly optimized under a reconstruction objective, encouraging latent actions to capture the temporal dynamics between consecutive observations. During the distillation stage, a latent policy is trained to predict latent actions produced by a frozen IDM, while in post-training, an action decoder is learned from limited real-world interaction data to map latent actions into executable physical actions.

Despite the effectiveness of this paradigm, existing works have primarily focused on improving the pretraining stage, emphasizing the learning of expressive latent action representations \cite{klepach2025object, nikulin2025latent, liu2026stamo, bu2026laof, xie2026flare, jeong2026learning} and their application to downstream such as vision-language-action (VLA) models \cite{ye2024latent, bjorck2025gr00t, bu2025agibot, bu2025learning}. In contrast, the distillation stage, serving as a critical bridge between representation learning and control, remains insufficiently explored, particularly in preserving the intrinsic structure of the pretrained latent action space. The core issue is that the IDM is trained on a relatively simple, often unimodal and non-causal inverse dynamics problem, whereas the latent policy must capture inherently multimodal future uncertainties. Prior work \cite{schmidt2023learning, chen2025villa, nikulin2025latent} typically relies on behavior cloning (BC) \cite{bain1995framework} to distill knowledge from the IDM into the latent policy, which tends to collapse inherently multimodal action distributions into unimodal ones. A natural alternative is flow matching (FM) \cite{lipman2022flow, liu2022flow, li2026back} for modeling multimodal action distributions, but naively applying it introduces a new challenge, as the stochasticity of the learned latent policy causes misalignment between latent actions and physical actions during post-training, degrading the effectiveness of the action decoder.

To address these, we propose Latent Action Flow Policy (LAFP), which leverages flow matching for the latent policy learning, preserving the well-structured representation learned in the pretraining phase. Additionally, in the post-training stage, we introduce an inference-time interpolation mechanism that performs flow matching inference by interpolating between the source Gaussian noise distribution and the target latent action distribution to train the action decoder, mitigating the misalignment between latent actions and physical actions. Our main contributions are as follows:

(1) The results on 16 PROCGEN tasks demonstrate that learning latent policies via flow matching can explicitly preserve the intrinsic structure of the pretrained latent action space, leading to consistent improvements in downstream imitation learning performance.

(2) we propose an inference time interpolation mechanism for training the action decoder under a flow-matching inference paradigm, enabling the resulting policy to achieve performance comparable to an action decoder directly trained with IDM supervision;

(3) Through ablation studies, we identify an optimal number of inference steps, which achieves an 8.2\% improvement in average success rate while incurring less than 1× additional training and inference overhead.

\section{Related work}
\label{relate_work}

\subsection{Latent action model}
Recent works have explored learning latent actions from videos through unsupervised or weakly supervised paradigms \cite{bu2026matters}. LAPO \cite{schmidt2023learning} introduces an unsupervised framework for latent action learning, while LAPA \cite{ye2024latent} extends this paradigm to VLA settings. To mitigate shortcut learning, CoMo \cite{yang2026learning} replaces the future observation input to the inverse dynamics model with the difference between future and current observations, discouraging direct encoding of future observations. UniVLA \cite{bu2025learning} combines DINOv2 \cite{oquab2023dinov2} features with language conditioning to learn task-centric latent actions, reducing the influence of task-irrelevant dynamics. Other works introduce explicit supervision to constrain latent actions. LAOM \cite{nikulin2025latent} uses limited action annotations to train an auxiliary action decoder during pretraining, improving representation quality and robustness to distractors. LAOF \cite{bu2026laof} further explores optical flow as a substitute for action supervision, enabling stable and improved training of LAOM under extremely limited action labels. VILLA-X \cite{chen2025villa} introduces robot-state and action prediction constraints, while SkeLAM \cite{lv2026skeleton} employs robot skeleton supervision to constrain a kinematic latent component. Studies on limited action supervision consistently show that even incorporating around 10\% labeled action data can significantly improve downstream imitation learning performance \cite{nikulin2025latent, zhang2025latent, bi2026motus}. We therefore adopt LAOM as our baseline, using the available action labels while avoiding potential errors from additional supervision signals.

\subsection{Generative policy learning}
Generative policy learning has emerged as a dominant paradigm in embodied intelligence due to its ability to model multimodal action distributions. Diffusion-based policies \cite{chi2025diffusion} formulate action generation as a conditional denoising process, enabling diverse and feasible behaviors under identical observations. However, their iterative denoising process results in high inference latency, limiting real time applicability. To address this issue, flow matching \cite{lipman2022flow, liu2022flow} based methods have been proposed as alternatives to diffusion models. These approaches directly learn continuous vector fields for state transitions, removing iterative denoising while maintaining high quality generation and significantly improving inference efficiency. For example, $\pi_0$ \cite{black2024pi_0} adopts flow matching as its policy head and ManiFlow \cite{yan2025maniflow} combines flow matching with consistency training to achieve high quality actions in 1–2 inference steps. More recently, Just Image Transformer (JiT) \cite{li2026back} revisits generative modeling from an x-prediction perspective, showing that directly predicting clean data improves stability and distributional fidelity in high dimensional generation tasks. Motivated by these advances, our work investigates how different generative objectives, including latent action prediction and vector field prediction, affect downstream task performance under different latent action dimensions.

\section{Method}
\label{method}

\begin{figure}[t!]
    \centering
    \includegraphics[width=1.0\textwidth]{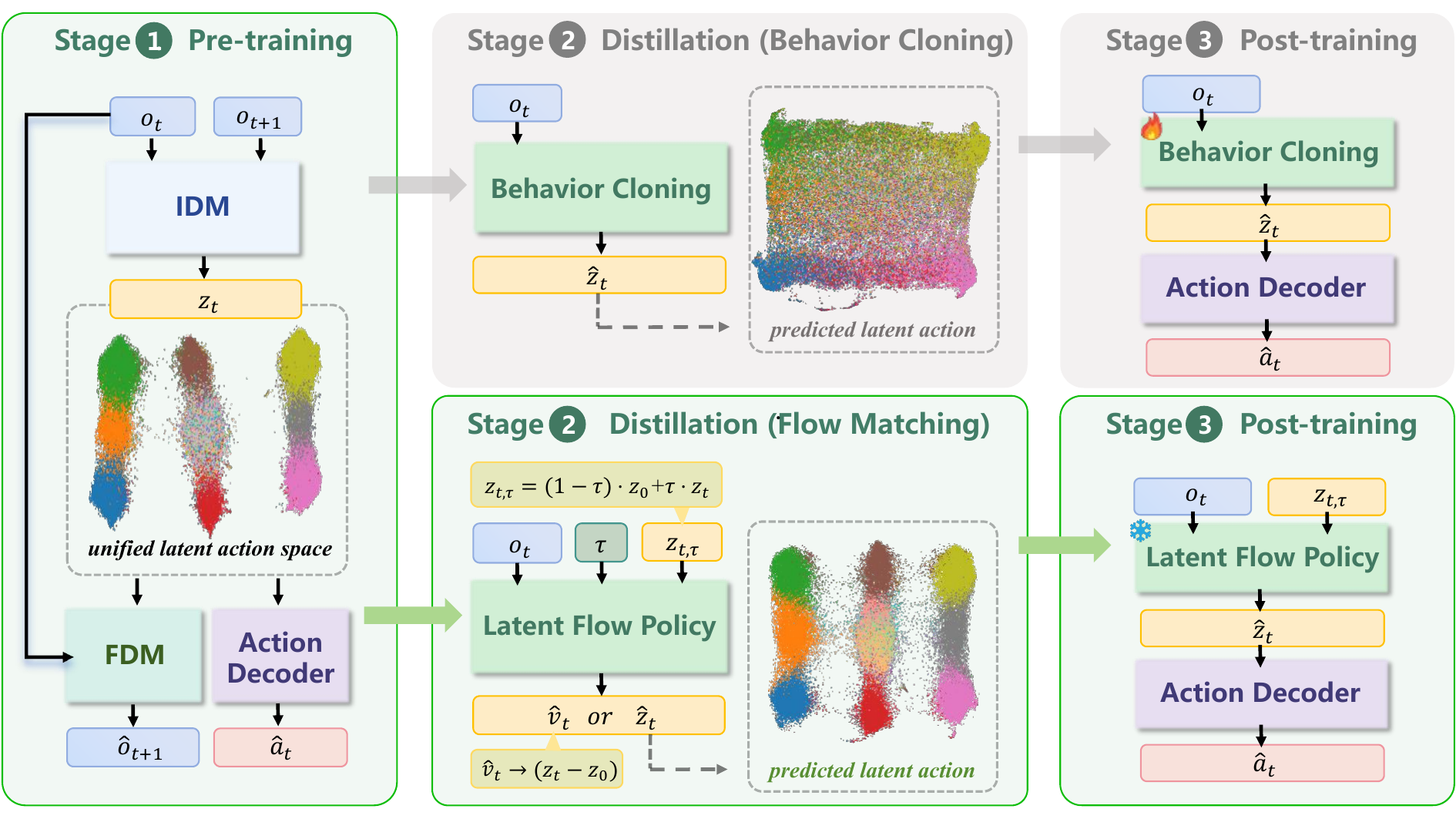}
    \caption{\textbf{Overview of LAFP.} Stage~1 learns an IDM-induced latent action space with an auxiliary action decoder. Stage~2 distills the frozen IDM targets using conditional flow matching. Stage~3 freezes the flow policy and trains the Post-AD with decoder aligned interpolation to reduce the mismatch between stochastic latent samples and physical action labels.}
    \label{fig:mainmethod}
\end{figure}

Figure~\ref{fig:mainmethod} presents Latent Action Flow Policy (LAFP) as a sequential three-stage pipeline. The latent action space learned from observation transitions in Stage~1 provides the targets for policy distillation in Stage~2, which learns to generate latent actions in this space from the current observation alone. In Stage~3, the flow policy remains fixed while a decoder is trained with limited action labels to map its latent outputs to executable actions. In this way, each stage builds on the preceding one while preserving the latent structure transferred through the pipeline.

\subsection{Latent Action Pre-training}
We follow LAOM \cite{nikulin2025latent} and jointly train an inverse dynamics model (IDM), a forward dynamics model (FDM), and an auxiliary action decoder (IDM-AD), as shown in Fig.~\ref{fig:mainmethod}. For an observation transition $(o_t,o_{t+1})$, where $t$ denotes the environment time step, the IDM produces a latent action $z_t$, and the FDM reconstructs $o_{t+1}$ from $(o_t,z_t)$. A VQ-VAE codebook \cite{van2017neural}, updated using exponential moving averages, discretizes the latent space. Let $e_t$ denote the codebook entry nearest to $z_t$. The representation objectives are
\begin{equation}
\mathcal{L}_{\mathrm{VQ}}=\lVert z_t-e_t\rVert_2^2,
\qquad
\mathcal{L}_{\mathrm{rec}}=\lVert \hat{o}_{t+1}-o_{t+1}\rVert_2^2.
\end{equation}
Approximately 10\% of the transitions include a physical action label $a_t$. The IDM-AD maps $z_t$ to $a_t$, and its supervised loss $\mathcal{L}_{\mathrm{IDM\text{-}AD}}$ is evaluated only on these labeled samples, whereas $\mathcal{L}_{\mathrm{VQ}}$ and $\mathcal{L}_{\mathrm{rec}}$ use the entire minibatch. Following LAOM, we separately set the weight of the supervised IDM-AD loss to $\lambda_{\mathrm{AD}}=0.1$. The complete objective is
\begin{equation}
\mathcal{L}_{\mathrm{stage1}}=\mathcal{L}_{\mathrm{VQ}}+\mathcal{L}_{\mathrm{rec}}+\lambda_{\mathrm{AD}}\mathcal{L}_{\mathrm{IDM\text{-}AD}}.
\end{equation}
The IDM-AD provides semantic regularization by encouraging latent actions to remain distinguishable by physical action class. Its action recovery accuracy therefore serves as a measure of the separability of the learned latent space. After latent action pre-training, the IDM, FDM, codebook, and IDM-AD are frozen.

\subsection{Flow Matching Policy Distillation}
The frozen IDM provides a latent target $z_t$ for each observation transition. A behavior cloning policy trained with a mean squared error (MSE) loss estimates the conditional mean $\mathbb{E}[z_t\mid o_t]$. When $p(z_t\mid o_t)$ is multimodal, this objective can merge distinct action modes into a single prediction. LAFP instead models the conditional latent action distribution with a flow policy $\pi_{\mathrm{flow}}$.

\begin{figure}[t!]
    \centering
    \includegraphics[width=1.0\textwidth]{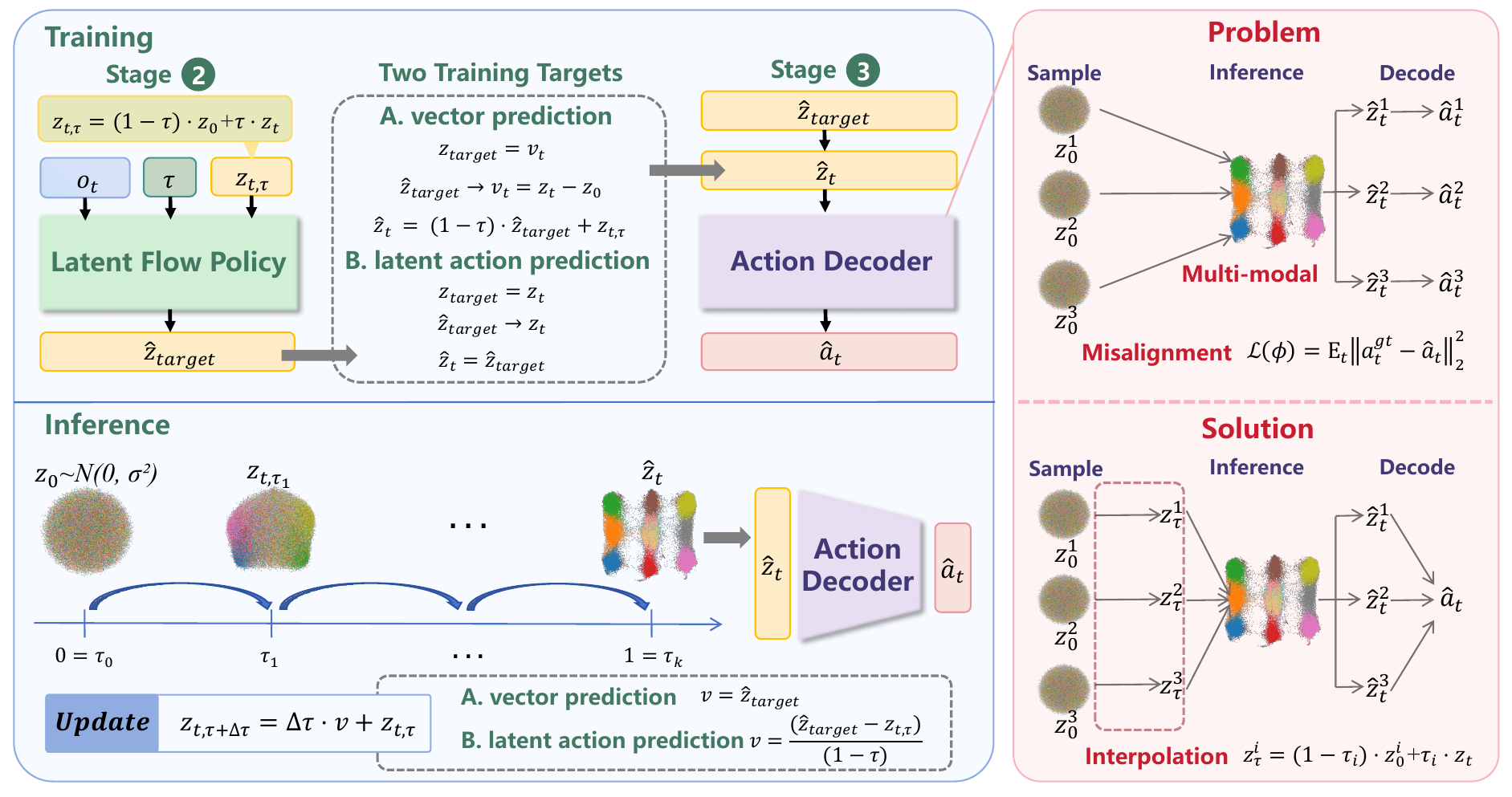}
    \caption{\textbf{Flow policy training and decoding.} \textit{Top:} x-prediction and v-prediction parameterize the same linear probability path. \textit{Right:} Direct sampling can pair different plausible latents with the same labeled action during Post-AD training; decoder aligned interpolation keeps the generated latent near its transition specific IDM target. \textit{Bottom:} At deployment, the frozen flow policy transports Gaussian noise to a latent action, which the Post-AD maps to an environment action.}
    \label{fig:method}
\end{figure}

As shown in Fig.~\ref{fig:method}, for each training transition, we draw a source latent $z_0\sim\mathcal{N}(0,I)$ and a flow time $\tau\sim\mathcal{U}[0,1]$, then construct the linear probability path
\begin{equation}
z_t^\tau=(1-\tau)z_0+\tau z_t.
\end{equation}
Here, $z_t^\tau$ denotes the latent state at flow time $\tau$ for the environment transition indexed by $t$. The policy receives $(z_t^\tau,\tau,o_t)$. We implement $\pi_{\mathrm{flow}}$ as a three-layer MLP, using a lightweight CNN to encode $o_t$ and sinusoidal embeddings to encode $\tau$. We consider two output parameterizations of the flow policy, distinguished by their prediction targets. The x-prediction model directly predicts the target latent $z_t$,
\begin{equation}
\hat z_t=\pi_{\mathrm{flow}}(z_t^\tau,\tau,o_t),
\qquad
\mathcal{L}_{x}=\mathbb{E}\lVert z_t-\hat z_t\rVert_2^2,
\end{equation}
whereas the v-prediction model predicts the displacement $v_t=z_t-z_0$ with $\mathcal{L}_{v}=\mathbb{E}\lVert v_t-\hat v_t\rVert_2^2$. Starting from Gaussian noise, Euler integration uses
\begin{equation}
u_{\theta}(z_t^\tau,\tau,o_t)=
\begin{cases}
(\hat z_t-z_t^\tau)/(1-\tau), & \text{x-prediction},\\
\hat v_t, & \text{v-prediction},
\end{cases}
\end{equation}
followed by $z_t^{\tau+\Delta\tau}=z_t^\tau+\Delta\tau\,u_{\theta}$. Unless stated otherwise, LAFP uses x-prediction with three integration steps, which provides the best balance between stability and efficiency in our ablations.

\subsection{Decoder Aligned Action Mapping}
We train an action decoder (Post-AD) $g_{\phi}$ to map latent actions generated by the flow policy to physical actions. The main challenge arises from the stochasticity of the flow policy. While a deterministic behavior cloning policy yields a single latent--action pair for each labeled transition, different samples of $z_0$ can produce different plausible latent actions $\hat z_t$ for the same observation $o_t$, whereas the dataset provides only one recorded action $a_t$. Pairing every sampled latent with this action can therefore assign the same label to latents with different action semantics. To prevent decoder training from altering the latent distribution learned during policy distillation, we freeze $\pi_{\mathrm{flow}}$ and optimize only $g_{\phi}$ using the supervised decoding loss $\mathcal{L}_{\mathrm{Post\text{-}AD}}$.
                                                 
One way to avoid this mismatch is to train $g_{\phi}$ directly on the IDM target pairs $(z_t,a_t)$, which assumes that the distilled policy remains aligned with the IDM latent space. Reusing the frozen IDM-AD without further training provides a stricter test of the same assumption. For comparison, Direct-Sample AD starts from Gaussian noise, applies the frozen flow solver, and trains $g_{\phi}$ on the resulting pairs $(\hat z_t,a_t)$. This exposes the decoder to policy generated latents but does not resolve the sampling induced label mismatch.

To retain policy generated decoder inputs while reducing this mismatch, LAFP uses decoder aligned interpolation. During Post-AD training, the IDM target $z_t$ is available for each labeled transition. We therefore initialize flow inference from an intermediate point between $z_0$ and $z_t$, rather than from pure Gaussian noise, and then apply the frozen flow solver. The resulting latent remains policy generated but is constrained toward its transition specific IDM target, providing a more consistent correspondence with $a_t$. Because only the decoder training inputs are modified, the interpolation neither updates the flow policy nor changes its learned latent distribution. At deployment, the IDM target is unavailable, so inference starts from Gaussian noise and the trained Post-AD decodes the generated latent action.

\begin{figure}[t!]
    \centering
    \includegraphics[width=0.9\textwidth]{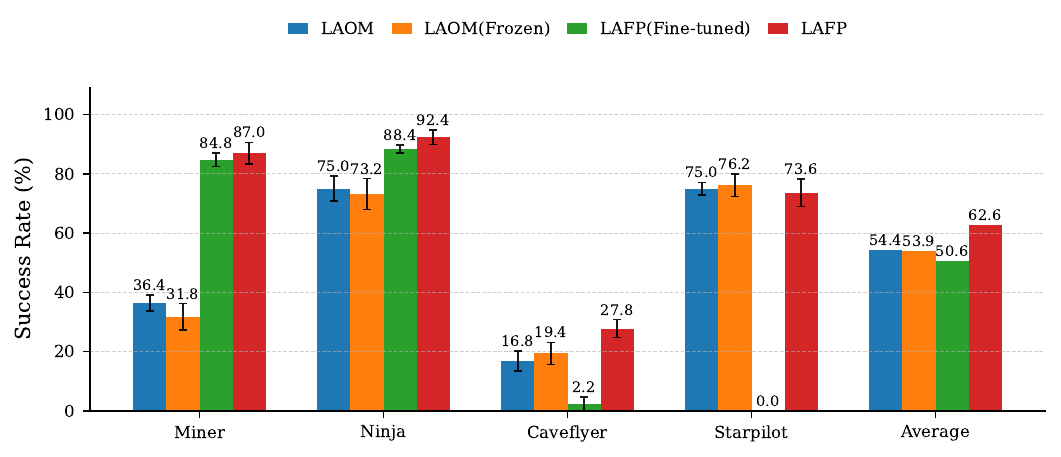}
    \caption{\textbf{Success rate comparison between LAOM and LAFP.} Results are averaged over 100 evaluation episodes across 5 random seeds and reported as mean $\pm$ standard deviation. Four representative environments are shown, corresponding to different behavioral categories: navigation (\textit{CaveFlyer}), platforming (\textit{Ninja}), collection (\textit{Miner}), and combat (\textit{StarPilot}). Full results on all 16 environments are provided in the Appendix. The rightmost panel reports the average success rate across all 16 environments, consistent with the following figures and tables.}
    \label{fig:baselines sr}
\end{figure}


\section{Experiments}
\label{experiment}
Our experiments address four questions: \textbf{(Q1)} Does the distilled policy remain aligned with the IDM latent action distribution? \textbf{(Q2)} Is stronger alignment associated with better downstream imitation performance? \textbf{(Q3)} Does decoder-aligned interpolation resolve the sampling-induced mismatch in Stage~3? \textbf{(Q4)} How do the prediction target and number of flow steps affect accuracy and efficiency?

\subsection{Experimental setup}
\label{experiment set}

\paragraph{Benchmark.} We evaluate our method on all 16 PROCGEN environments \cite{cobbe2020leveraging}. PROCGEN provides procedurally generated training and testing levels with substantial visual and structural variation, and covers diverse task categories, including navigation, platforming, collection, and combat. This diversity provides a broad test bed for evaluating the effectiveness and generality of our method across different task categories. Moreover, the discrete action spaces provide explicit action classes, making latent action separability, distributional alignment, and decoder transfer directly measurable. Since the environments differ in observations, dynamics, and action semantics, we train a separate policy for each environment. For each environment, we collect approximately 2.6M training frames and 6.5K held-out frames using an expert PPO policy \cite{schulman2017proximal} trained for 50M time steps. Latent action pre-training, latent policy distillation, and action decoder post-training run for 50K, 120K, and 5K optimization steps, respectively. The behavior cloning distillation baseline converges after 60K steps; extending its training to 120K yields negligible improvement. Unless stated otherwise, the latent action dimension is 128 and LAFP uses three flow steps for inference.
\paragraph{Baselines and variants.} Our primary baseline is LAOM, which uses the same pre-training VQ-VAE latent space \cite{van2017neural}, including IDM-AD regularization, but distills the latent policy by behavior cloning. Both methods use the same 10\% action-labeled data subset in latent action pre-training and action decoder post-training. We evaluate two post-training variants:

\begin{figure}[t!]
    \centering
    \fbox{%
        \begin{minipage}{\dimexpr0.95\textwidth-2\fboxsep-2\fboxrule\relax}
            \centering
            \includegraphics[width=\textwidth]{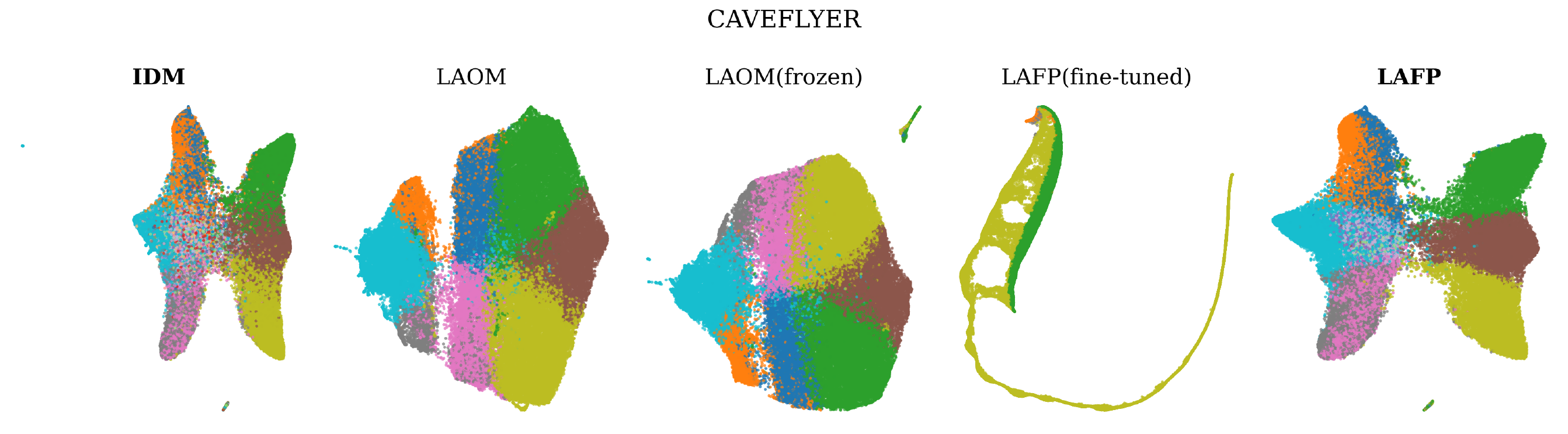}\\[2pt]
            \includegraphics[width=\textwidth]{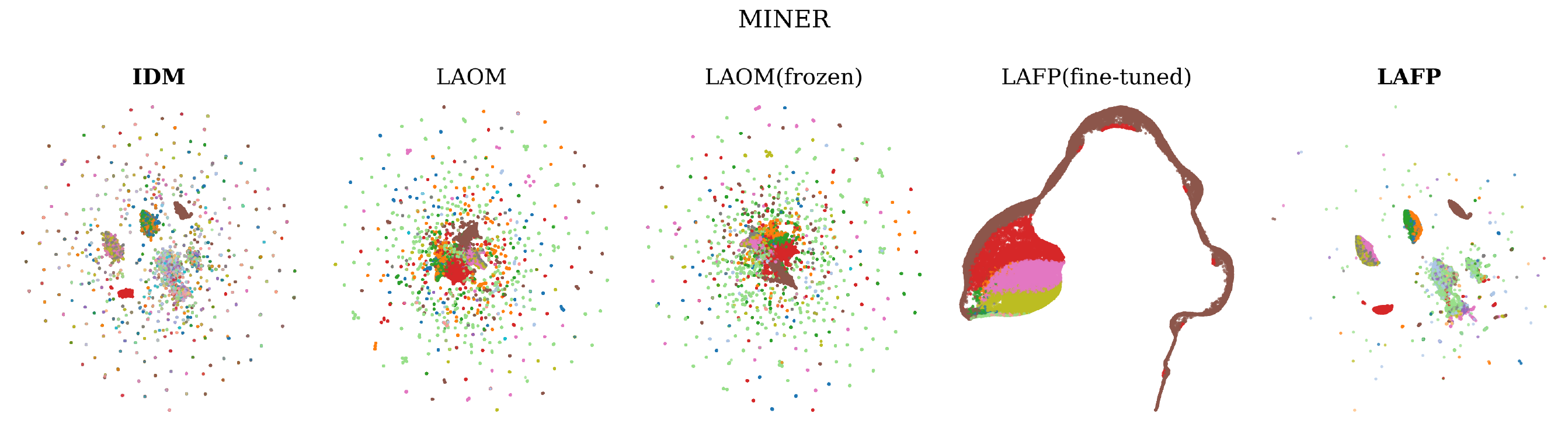}\\[2pt]
            \includegraphics[width=\textwidth]{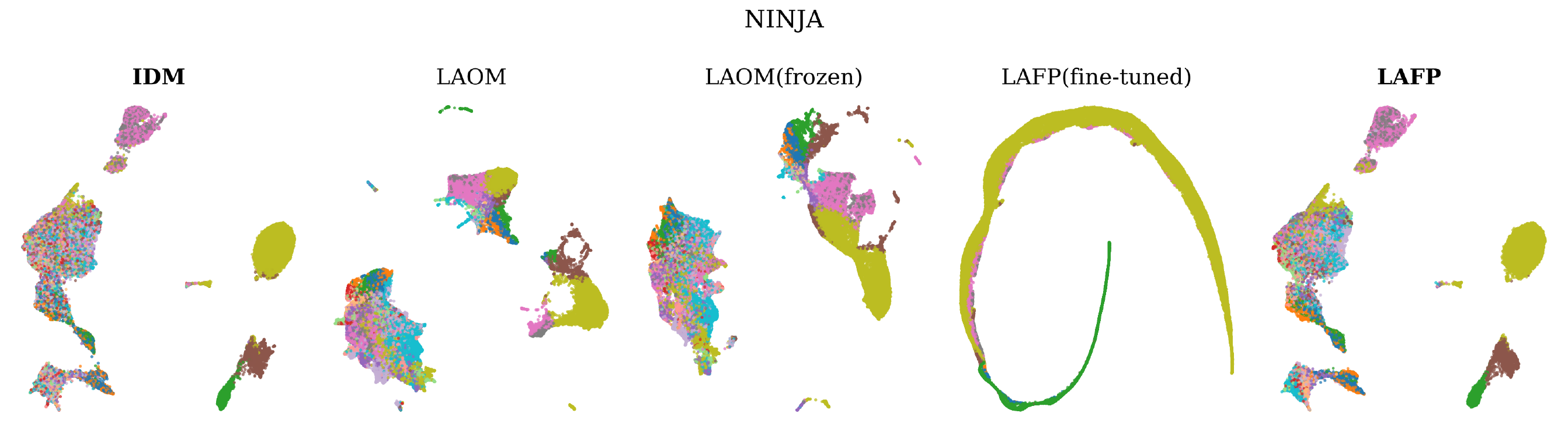}\\[2pt]
            \includegraphics[width=\textwidth]{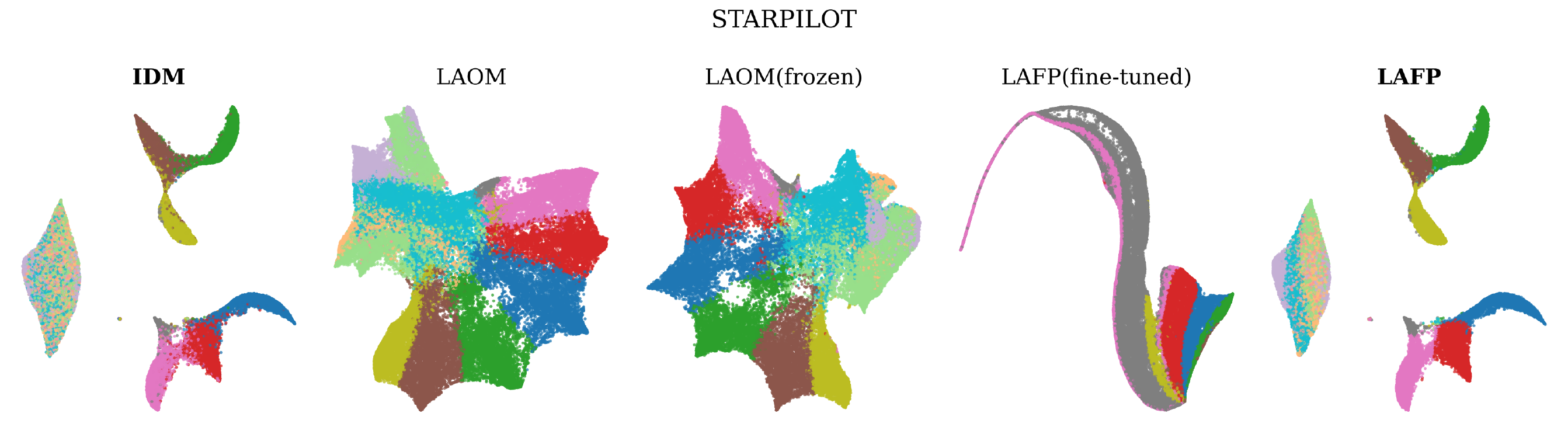}\\[2pt]
            \includegraphics[width=0.5\textwidth]
            {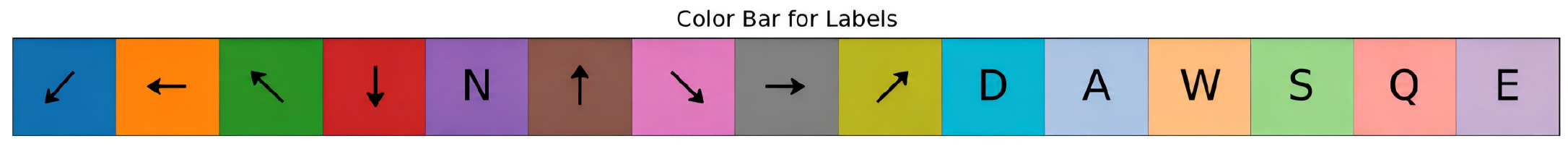}
    \end{minipage}} 
    \caption{\textbf{UMAP projections of latent action spaces for the IDM and downstream policies across four representative environments.} Each row corresponds to one environment, while columns are ordered as the IDM, LAOM, LAOM(Frozen), LAFP(Fine-tuned) and LAFP latent space for reference. The LAFP and IDM columns are highlighted to facilitate direct comparison. Colors denote discrete action classes: the IDM latent space is colored using ground-truth action labels, while downstream policies are colored using decoded action labels.}
    \label{fig:IDM bc flow umap}
\end{figure}

\begin{itemize}
    \item \textbf{Fine-tuned}: the distilled latent policy and the Post-AD are updated jointly in action decoder post-training stage, allowing the latent representation to adapt to the downstream action space.
    \item \textbf{Frozen}: the distilled latent policy is fixed and only the Post-AD is trained, preserving the latent structure learned during distillation.
\end{itemize}

Accordingly, \textbf{LAOM} and \textbf{LAOM(Frozen)} are behavior cloning baselines, while \textbf{LAFP} denotes our complete method with a frozen latent policy and decoder aligned interpolation. We additionally include \textbf{LAFP(Fine-tuned)} and \textbf{Direct-Sample AD} as comparison variants to isolate the effects of policy freezing and decoder aligned interpolation, respectively.

\paragraph{Metrics.} \emph{IDM alignment} compares a frozen policy's latent distribution $P$ with the IDM reference $Q$ on the same held-out observations using squared maximum mean discrepancy (MMD),
\begin{equation}
\operatorname{MMD}^{2}(P,Q)=\mathbb{E}k(x,x')+\mathbb{E}k(y,y')-2\mathbb{E}k(x,y),
\end{equation}
where $x,x'\sim P$, $y,y'\sim Q$, and $k$ is the Gaussian RBF kernel \(k(x,y)=\exp\left(-\frac{\|x-y\|_2^2}{2\sigma^2}\right)\). Lower MMD means closer alignment to the IDM distribution. For downstream evaluation, \emph{success rate} is the fraction of episodes whose cumulative reward reaches the environment specific maximum defined by Cobbe et al. \cite{cobbe2020leveraging}. Each method is evaluated over five random seeds, with 100 episodes run for each seed; we report the mean and standard deviation across seeds. Full per-environment MMD and success rate results appear in the Appendix.

\begin{table}[t!]
  \caption{Impact of IDM-AD regularization on pretrained latent action quality and downstream
  policy performance. IDM label recovery is the IDM-AD accuracy on held-out IDM latent actions,
  and $\Delta\mathrm{Acc}$ denotes the gain in this accuracy produced by IDM-AD regularization.
  $\Delta\mathrm{Score}$ denotes the downstream policy gain (AD $-$ NAD).
  The largest policy improvement is highlighted in bold with a light red background.}
  \vspace{5pt}
  \label{tab:ad_analysis}
  \centering
  \small
  \renewcommand{\arraystretch}{0.50}
  \begin{tabular}{lccccc}
    \toprule
    Environment & $\Delta\mathrm{Acc}$ & LAOM & LAOM (Frozen) & LAFP (Fine-tuned) & LAFP \\
    \midrule
    CaveFlyer & 13.3 & $+$3.6 & $+$6.2 & $-$7.6 & \cellcolor{red!12}\textbf{$+$10.6} \\
    Miner     & 7.5  & $-$11.8 & $-$9.2 & $+$6.2 & \cellcolor{red!12}\textbf{$+$14.2} \\
    Ninja     & 12.3 & $-$4.8 & $-$4.2 & $+$10.4 & \cellcolor{red!12}\textbf{$+$11.4} \\
    StarPilot & 5.0  & $+$0.4 & $-$1.0 & $-$0.2 & \cellcolor{red!12}\textbf{$+$4.8} \\
    \midrule
    Average   & 7.8  & $+$1.2 & $+$6.4 & $+$7.8 & \cellcolor{red!18}\textbf{$+$10.3} \\
    \bottomrule
  \end{tabular}
\end{table}

\subsection{Latent Action Structure and Downstream Performance}

\paragraph{Latent action structure after policy distillation (Q1).}
We first examine whether the distilled policy retains the latent action structure learned by the IDM. The UMAP visualizations in Fig.~\ref{fig:IDM bc flow umap} show that LAFP more closely preserves the IDM cluster structure, whereas LAOM(Frozen) shows larger deviations from the IDM reference across the four representative environments. Since UMAP provides only a qualitative two-dimensional projection, we additionally quantify distributional alignment in the full latent space using squared MMD. As shown in Table~\ref{tab:mmd-main}, LAFP obtains substantially lower MMD than LAOM(Frozen) in all four environments. The same trend holds across all 16 environments, with average MMD values of 0.00060 and 0.03846, respectively. To examine whether this alignment also preserves the action semantics encoded by the IDM, we replace each frozen policy's Post-AD with the pretrained IDM-AD without retraining the decoder. This replacement leaves LAFP's average success rate unchanged at 62.6\%, but reduces LAOM(Frozen) from 53.9\% to 44.6\% (Table~\ref{tab:ad_compare}). Since the IDM-AD is trained only on IDM latents, its transfer performance indicates that LAFP remains more compatible with the action semantics of the IDM-induced latent space. The combined evidence from UMAP, MMD, and decoder transfer shows that flow matching policy distillation more faithfully preserves the pretrained latent action structure.

\begin{table}[t!]
  \caption{Squared MMD between the distilled-policy latent distribution and the IDM reference in four representative PROCGEN environments. Lower values indicate better structure preservation. Results are averaged over five random seeds; complete results for all 16 environments are reported in Table~\ref{tab:mmd-results}.}
  \label{tab:mmd-main}
  \centering
  \small
  \begin{tabular}{lcc}
    \toprule
    Environment & LAFP & LAOM(Frozen) \\
    \midrule
    CaveFlyer & \textbf{0.00045} & 0.04481 \\
    Ninja     & \textbf{0.00047} & 0.01473 \\
    Miner     & \textbf{0.00070} & 0.08450 \\
    StarPilot & \textbf{0.00068} & 0.00679 \\
    \bottomrule
  \end{tabular}
\end{table}

\paragraph{Latent space alignment and downstream performance.}
We next investigate how alignment with the IDM latent distribution relates to downstream performance. Holding the environment and pretrained IDM latent space fixed, LAFP achieves lower MMD than LAOM(Frozen) in all 16 environments and an equal or higher success rate in 13 of them. Its average success rate is 62.6\%, compared with 53.9\% for LAOM(Frozen) and 54.4\% for the fine-tuned LAOM baseline (Figure~\ref{fig:baselines sr}). The matched comparison with LAOM(Frozen) controls for policy updates during action decoder post-training and shows that LAFP combines closer alignment with the IDM distribution with higher downstream success. The fine-tuned variants are analyzed separately below since their latent spaces are further modified during post-training.

\paragraph{Fine-tuned versus frozen latent policies.}
The effect of fine-tuning differs between the two policy families. For LAFP, jointly updating the flow matching policy and Post-AD during action decoder post-training reduces the average success rate from 62.6\% to 50.6\%. Performance declines moderately in several environments and falls to zero in Dodgeball and StarPilot. Both tasks are sensitive to individual action errors and offer few alternative valid actions, leaving limited room for recovery once fine-tuning perturbs the pretrained latent geometry. Because flow matching learns a transport field over this geometry, such perturbations can distort the field and degrade performance. In contrast, fine-tuning changes LAOM only slightly, from 53.9\% to 54.4\%, because its pointwise mapping can adapt directly to the labeled action objective. This comparison shows that freezing the flow policy during action decoder post-training is important for retaining the structure learned during policy distillation.

\subsection{Ablation Studies}

\begin{table}[t!]
  \caption{Impact of decoder training and transfer on LAOM(Frozen) and LAFP. IDM-AD denotes the pretrained decoder reused without further optimization. Direct-Sample AD trains the Post-AD on latents generated from Gaussian noise using three flow steps, without decoder-aligned interpolation. The remaining columns use a Post-AD trained after policy distillation.}
  \vspace{5pt}
  \label{tab:ad_compare}
  \centering
  \resizebox{\textwidth}{!}{%
  \small
  \renewcommand{\arraystretch}{0.95}
  \begin{tabular}{lccccc}
    \toprule
    Environment & LAOM(Frozen) IDM-AD & LAOM(Frozen) & LAFP Direct-Sample AD & LAFP IDM-AD & LAFP\\
    \midrule
    Miner     & 0.2${\pm}$3.9  & 31.8${\pm}$5.2 & 86.2${\pm}$2.6 & 81.4${\pm}$0.4 & \cellcolor{red!12}\textbf{87.0${\pm}$2.5} \\
    Ninja     & 63.8${\pm}$2.1 & 73.2${\pm}$3.7 & 92.4${\pm}$1.3 & \cellcolor{red!12}\textbf{93.0${\pm}$6.7} & 92.4${\pm}$3.0 \\
    CaveFlyer & 19.4${\pm}$5.3 & 19.4${\pm}$4.4 & 18.4${\pm}$4.4 & \cellcolor{red!12}\textbf{33.0${\pm}$7.2} & 27.8${\pm}$3.6 \\
    StarPilot & 70.6${\pm}$2.7 & 76.2${\pm}$3.8 & 73.8${\pm}$4.5 & \cellcolor{red!12}\textbf{76.6${\pm}$0.9} & 73.6${\pm}$4.6 \\
    \midrule
    Average   & 44.6 & 53.9 & 61.3 & \cellcolor{red!12}\textbf{62.6} & \cellcolor{red!12}\textbf{62.6} \\
    \bottomrule
  \end{tabular}%
  }
\end{table}

\paragraph{IDM-AD regularization.}
Table~\ref{tab:ad_analysis} evaluates the effect of IDM-AD regularization during latent action pretraining. It improves IDM label recovery accuracy in all four representative environments, with an average $\Delta\mathrm{Acc}$ of $+7.8$ points across all 16 environments. For LAFP, this improvement is accompanied by an average downstream gain of $10.3$ points, compared with $1.2$ points for LAOM. Since LAFP distills the latent action distribution, clearer separation among action modes provides a better structured target for the flow matching policy. These results provide complementary evidence that latent action quality is important for downstream flow matching policy learning.

\paragraph{Decoder aligned interpolation.}
To isolate the effect of interpolation, Direct-Sample AD uses the same frozen flow policy but trains the Post-AD on latents sampled directly from Gaussian noise. It achieves an average success rate of 61.3\%, compared with 62.6\% for LAFP. The difference is larger in CaveFlyer, where success increases from 18.4\% to 27.8\% with interpolation (Table~\ref{tab:ad_compare}). During decoder training, each generated latent is interpolated toward the transition specific IDM latent, maintaining consistency between the decoder input and its ground-truth action label without updating the frozen flow policy. The 9.4-point improvement in CaveFlyer, despite a 1.3-point average gain, suggests that interpolation is particularly beneficial in tasks with more multimodal action distributions, where stochastic latent samples are more likely to differ from the single action label available for decoder training.

\begin{figure}[t!]
    \centering
    \includegraphics[width=1.0\textwidth]{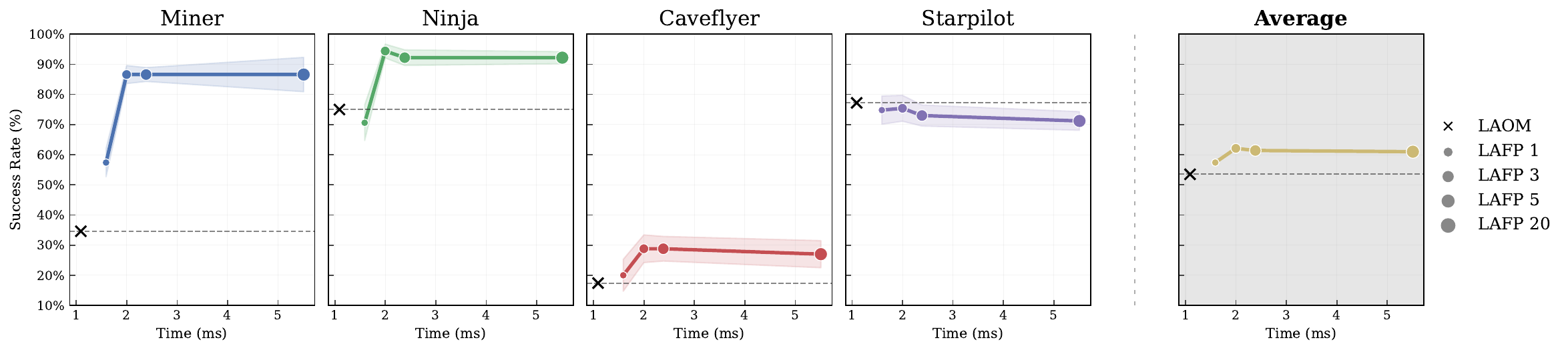}
    \caption{Performance and per-action inference time of flow matching under different inference step budgets, compared with behavior cloning. LAFP-$k$ denotes a flow matching policy using $k$ inference steps.}
    \label{fig:time-performance}
\end{figure}

\begin{figure}[t!]
    \centering
    \includegraphics[width=1.0\textwidth]{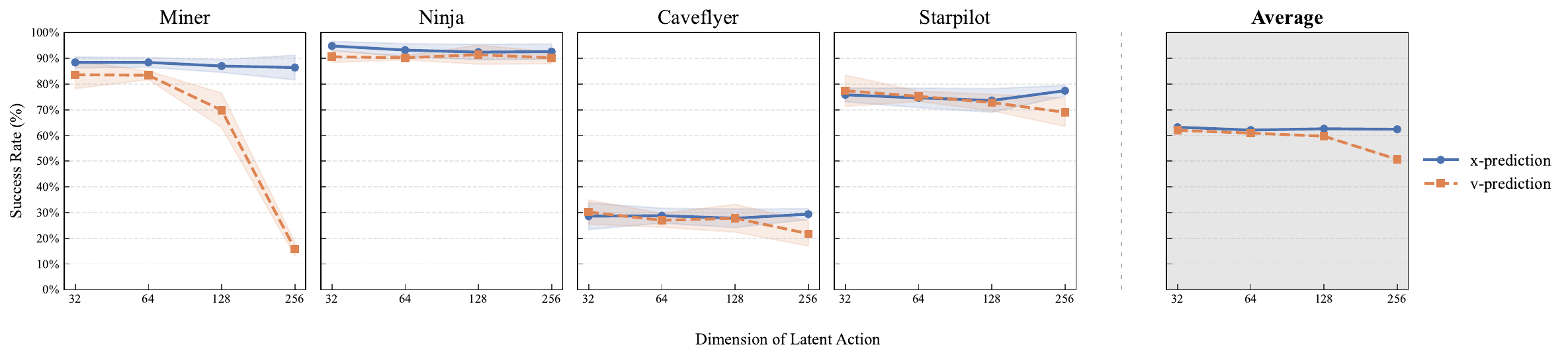}
    \caption{Comparison between latent action prediction (x-prediction) and vector field prediction (v-prediction) under different latent action dimensions.}
    \label{fig:pred-performance}
\end{figure}

\paragraph{Inference steps budget.}
Figure~\ref{fig:time-performance} compares performance and latency under different flow steps budgets. Performance improves with the first few steps and then saturates, while inference time increases monotonically. Three steps provide the best trade-off and are therefore used as the default setting.
Additional steps initially improve the numerical integration of the learned flow, but provide little benefit once the trajectory is sufficiently resolved.

\paragraph{Prediction target and latent dimensionality.}
Figure~\ref{fig:pred-performance} compares latent action prediction (x-prediction) with vector field prediction (v-prediction) across latent dimensions. X-prediction remains stable across all dimensions and performs better at dimensions 32--128, whereas v-prediction becomes less stable as dimensionality increases and shows the largest degradation at dimension 256. This trend suggests that direct supervision of the target latent is more stable as dimensionality grows, whereas errors in the predicted displacement can accumulate during integration.

\section{Discussion}
Our results indicate that preserving the IDM-induced latent structure during policy distillation and action decoder post-training is closely associated with downstream performance. LAFP achieves lower MMD to the IDM reference in every environment, allows the IDM-AD to be reused without an average performance loss, and performs best when the flow matching policy remains frozen during Post-AD training. Together, these results show that downstream decoding benefits when the distilled latents retain the distributional structure and action semantics of the pretrained IDM latent space. Flow matching supports this preservation by modeling the conditional latent distribution, while decoder aligned interpolation reduces the mismatch between stochastic latent samples and their unique action labels for each transition during Post-AD training. With three inference steps, LAFP provides a favorable balance between performance and efficiency, adding less than $1\times$ inference overhead relative to behavior cloning.

\section{Limitations and future work}
\label{limit}
This study focuses on offline imitation learning in discrete PROCGEN environments, with a separate model trained for each game. Its effectiveness in continuous control, multi-task, and real world settings therefore remains to be evaluated. Moreover, jointly fine-tuning the flow policy and Post-AD is unstable in some tasks, motivating optimization strategies that adapt both components without disrupting the learned latent geometry. LAFP requires multiple flow inference steps and therefore has a higher computational cost than one-step policies, although the default three-step configuration introduces only a limited increase in latency. Future work will address these limitations through one-step flow distillation, structure preserving joint optimization, shared latent policies across tasks, and evaluation in continuous control and real world settings.

\section*{Acknowledgment}
We thank Shanghai Institute for Mathematics and Interdisciplinary Sciences (SIMIS) for their financial support. This work was funded by SIMIS under Grant No. SIMIS-ID2025-RB. The authors are grateful for the resources and facilities provided by SIMIS, which were essential for the completion of this work.

\bibliographystyle{ieeenat_fullname}
\bibliography{references}

@String(CVPR= {IEEE Conf. Comput. Vis. Pattern Recog.})

@String(ECCV= {Eur. Conf. Comput. Vis.})

@String(NIPS= {Adv. Neural Inform. Process. Syst.})

@String(ICLR = {Int. Conf. Learn. Represent.})

@String(CVPR  = {CVPR})

@String(ECCV  = {ECCV})

@String(NIPS  = {NeurIPS})

@String(ICLR  = {ICLR})

@inproceedings{cobbe2020leveraging,
	title={Leveraging procedural generation to benchmark reinforcement learning},
	author={Cobbe, Karl and Hesse, Chris and Hilton, Jacob and Schulman, John},
	booktitle={ICML},
	year={2020}
}

@article{schulman2017proximal,
  title={Proximal policy optimization algorithms},
  author={Schulman, John and Wolski, Filip and Dhariwal, Prafulla and Radford, Alec and Klimov, Oleg},
  journal={arXiv preprint arXiv:1707.06347},
  year={2017}
}

@article{bu2026matters,
  title={What matters for latent actions in robot learning},
  author={Bu, Xizhou and Hu, Qingda and Zhou, Lei and Zhang, Lingfeng and Tang, Yingbo and Liu, Zihao and Tao, Xinyi and Ma, Zhiqiang and Huang, Qingqiu and Tang, Chufeng and others},
  journal={arXiv preprint arXiv:2608.19613},
  year={2026}
}

@inproceedings{van2017neural,
	title={Neural discrete representation learning},
	author={Van Den Oord, Aaron and Vinyals, Oriol and Koray Kavukcuoglu},
	booktitle=NIPS,
	volume={30},
	year={2017}
}

@article{lv2026skeleton,
  title={Skeleton-Grounded Latent Action Model for Robot World Models},
  author={Lv, Qi and Lin, Jiayi and Jiang, Yuxin and Wang, Jianzong and Deng, Xiang and Zheng, Mike Shou},
  journal={arXiv preprint arXiv:2610.31474},
  year={2026}
}

@article{yu2026latent,
  title={The latent space: Foundation, evolution, mechanism, ability, and outlook},
  author={Yu, Xinlei and Chen, Zhangquan and He, Yongbo and Fu, Tianyu and Yang, Cheng and Xu, Chengming and Ma, Yue and Hu, Xiaobin and Cao, Zhe and Xu, Jie and others},
  journal={arXiv preprint arXiv:2604.02029},
  year={2026}
}

@article{mccarthy2025towards,
	title={Towards generalist robot learning from internet video: A survey},
	author={McCarthy, Robert and Tan, Daniel CH and Schmidt, Dominik and Acero, Fernando and Herr, Nathan and Du, Yilun and Thuruthel, Thomas G and Li, Zhibin},
	journal={Journal of Artificial Intelligence Research},
	volume={83},
	year={2025}
}

@inproceedings{nie2026lary,
  title={{LARY}: A Latent Action Representation Yielding Benchmark},
  author={Nie, Dujun and Chen, Fengjiao and Lv, Qi and Kuang, Jun and Li, Xiaoyu and Cao, Xuezhi},
  booktitle={ECCV},
  year={2026}
}

@article{chi2025diffusion,
  title={Diffusion policy: Visuomotor policy learning via action diffusion},
  author={Chi, Cheng and Xu, Zhenjia and Feng, Siyuan and Cousineau, Eric and Du, Yilun and Burchfiel, Benjamin and Tedrake, Russ and Song, Shuran},
  journal={The International Journal of Robotics Research},
  volume={44},
  number={10-11},
  pages={1684--1704},
  year={2025},
  publisher={Sage Publications Sage UK: London, England}
}

@article{lipman2022flow,
  title={Flow matching for generative modeling},
  author={Lipman, Yaron and Chen, Ricky TQ and Ben-Hamu, Heli and Nickel, Maximilian and Le, Matt},
  journal={arXiv preprint arXiv:2210.02747},
  year={2022}
}

@article{liu2022flow,
  title={Flow straight and fast: Learning to generate and transfer data with rectified flow},
  author={Liu, Xingchao and Gong, Chengyue and Liu, Qiang},
  journal={arXiv preprint arXiv:2209.03003},
  year={2022}
}

@inproceedings{li2026back,
  title={Back to basics: Let denoising generative models denoise},
  author={Li, Tianhong and He, Kaiming},
  booktitle={CVPR},
  year={2026}
}

@article{black2024pi_0,
  title={$\pi_0$: A Vision-Language-Action Flow Model for General Robot Control},
  author={Black, Kevin and Brown, Noah and Driess, Danny and Esmail, Adnan and Equi, Michael and Finn, Chelsea and Fusai, Niccolo and Groom, Lachy and Hausman, Karol and Ichter, Brian and others},
  journal={arXiv preprint arXiv:2410.24164},
  year={2024}
}

@inproceedings{bain1995framework,
  title={A Framework for Behavioural Cloning.},
  author={Bain, Michael and Sammut, Claude},
  booktitle={Machine intelligence 15},
  pages={103--129},
  year={1995}
}

@article{oquab2023dinov2,
  title={{DINOv2}: Learning robust visual features without supervision},
  author={Oquab, Maxime and Darcet, Timoth{\'e}e and Moutakanni, Th{\'e}o and Vo, Huy and Szafraniec, Marc and Khalidov, Vasil and Fernandez, Pierre and Haziza, Daniel and Massa, Francisco and El-Nouby, Alaaeldin and others},
  journal={arXiv preprint arXiv:2304.07193},
  year={2023}
}

@inproceedings{bi2026motus,
  title={Motus: A unified latent action world model},
  author={Bi, Hongzhe and Tan, Hengkai and Xie, Shenghao and Wang, Zeyuan and Huang, Shuhe and Liu, Haitian and Zhao, Ruowen and Feng, Yao and Xiang, Chendong and Rong, Yinze and others},
  booktitle={CVPR},
  year={2026}
}

@inproceedings{schmidt2023learning,
	title={Learning to act without actions},
	author={Schmidt, Dominik and Jiang, Minqi},
	booktitle=ICLR,
	year={2024}
}

@inproceedings{ye2024latent,
	title={Latent action pretraining from videos},
	author={Ye, Seonghyeon and Jang, Joel and Jeon, Byeongguk and Joo, Sejune and Yang, Jianwei and Peng, Baolin and Mandlekar, Ajay and Tan, Reuben and Chao, Yu-Wei and Lin, Bill Yuchen and others},
	booktitle=ICLR,
	year={2025}
}

@article{bjorck2025gr00t,
	title={{GR00T N1}: An open foundation model for generalist humanoid robots},
	author={Bjorck, Johan and Casta{\~n}eda, Fernando and Cherniadev, Nikita and Da, Xingye and Ding, Runyu and Fan, Linxi and Fang, Yu and Fox, Dieter and Hu, Fengyuan and Huang, Spencer and others},
	journal={arXiv preprint arXiv:2503.14734},
	year={2025}
}

@inproceedings{bu2025agibot,
	title={Agibot world colosseo: A large-scale manipulation platform for scalable and intelligent embodied systems},
	author={Bu, Qingwen and Cai, Jisong and Chen, Li and Cui, Xiuqi and Ding, Yan and Feng, Siyuan and Gao, Shenyuan and He, Xindong and Hu, Xuan and Huang, Xu and others},
	year={2025},
	booktitle={IROS}
}

@inproceedings{bu2025learning,
	title={Learning to Act Anywhere with Task-centric Latent Actions},
	author={Bu, Qingwen and Yang, Yanting and Cai, Jisong and Gao, Shenyuan and Ren, Guanghui and Yao, Maoqing and Luo, Ping and Li, Hongyang},
	year={2025},
	booktitle={RSS},
}

@article{yan2025maniflow,
  title={{ManiFlow}: A general robot manipulation policy via consistency flow training},
  author={Yan, Ge and Zhu, Jiyue and Deng, Yuquan and Yang, Shiqi and Qiu, Ri-Zhao and Cheng, Xuxin and Memmel, Marius and Krishna, Ranjay and Goyal, Ankit and Wang, Xiaolong and others},
  journal={arXiv preprint arXiv:2509.01819},
  year={2025}
}

@inproceedings{yang2026learning,
  title={{CoMo}: Learning continuous latent motion from internet videos for scalable robot learning},
  author={Yang, Jiange and Shi, Yansong and Zhu, Haoyi and Liu, Mingyu and Ma, Kaijing and Wang, Yating and Wu, Gangshan and He, Tong and Wang, Limin},
  booktitle={CVPR},
  year={2026}
}

@inproceedings{klepach2025object,
	title={Object-Centric Latent Action Learning},
	author={Klepach, Albina and Nikulin, Alexander and Zisman, Ilya and Tarasov, Denis and Derevyagin, Alexander and Polubarov, Andrei and Lyubaykin, Nikita and Kurenkov, Vladislav},
	booktitle={7th Robot Learning Workshop: Towards Robots with Human-Level Abilities},
	year={2025}
}

@inproceedings{zhang2025latent,
	title={What Do Latent Action Models Actually Learn?},
	author={Zhang, Chuheng and Pearce, Tim and Zhang, Pushi and Wang, Kaixin and Chen, Xiaoyu and Shen, Wei and Zhao, Li and Bian, Jiang},
	booktitle=NIPS,
	year={2025}
}

@inproceedings{nikulin2025latent,
	title={Latent action learning requires supervision in the presence of distractors},
	author={Nikulin, Alexander and Zisman, Ilya and Tarasov, Denis and Lyubaykin, Nikita and Polubarov, Andrei and Kiselev, Igor and Kurenkov, Vladislav},
	booktitle={ICML},
	year={2025}
}

@inproceedings{chen2025villa,
	title={{VILLA-X}: enhancing latent action modeling in vision-language-action models},
	author={Chen, Xiaoyu and Wei, Hangxing and Zhang, Pushi and Zhang, Chuheng and Wang, Kaixin and Guo, Yanjiang and Yang, Rushuai and Wang, Yucen and Xiao, Xinquan and Zhao, Li and others},
	booktitle=ICLR,
	year={2026}
}

@article{jeong2026learning,
  title={Learning to Act Robustly with View-Invariant Latent Actions},
  author={Jeong, Youngjoon and Chun, Junha and Kim, Taesup},
  journal={arXiv preprint arXiv:2601.02994},
  year={2026}
}

@inproceedings{liu2026stamo,
  title={{StaMo}: Unsupervised learning of generalizable robot motion from compact state representation},
  author={Liu, Mingyu and Shu, Jiuhe and Chen, Hui and Li, Zeju and Zhao, Canyu and Yang, Jiange and Gao, Shenyuan and Chen, Hao and Shen, Chunhua},
  booktitle={CVPR},
  year={2026}
}

@inproceedings{bu2026laof,
  title={{LAOF}: Robust latent action learning with optical flow constraints},
  author={Bu, Xizhou and Lyu, Jiexi and Sun, Fulei and Yang, Ruichen and Ma, Zhiqiang and Li, Wei},
  booktitle={CVPR},
  year={2026}
}

@article{xie2026flare,
  title={{FLARE}: Learning Future-Aware Latent Representations from Vision-Language Models for Autonomous Driving},
  author={Xie, Chengen and Sima, Chonghao and Li, Tianyu and Sun, Bin and Wu, Junjie and Hao, Zhihui and Li, Hongyang},
  journal={arXiv preprint arXiv:2601.05611},
  year={2026}
}


\clearpage
\appendix
\begin{center}
    \Large \textbf{Appendix}
\end{center}
\section*{Overview}

This appendix provides supplementary details and experimental results omitted from the main paper due to page constraints. It is organized as follows:

\begin{itemize}
  \item Sec.~\ref{appendix:full_results} reports success rates and MMD alignment scores across all 16 PROCGEN environments. Extended ablations cover IDM-AD regularization, Post-AD variants, the time--performance trade-off, and the flow prediction target.
  \item Sec.~\ref{appendix:umap_full} provides additional UMAP visualizations of latent action spaces for the remaining environments not shown in the main paper.
  \item Sec.~\ref{appendix:details} details the model architecture, normalization procedure, and computational cost of all methods.
\end{itemize}

\section{Full results on all PROCGEN environments}
\label{appendix:full_results}

In the main paper, we report results on four representative environments. To provide a comprehensive evaluation, we extend the comparison to all 16 PROCGEN environments.


\begin{table}[htbp]
  \caption{Performance comparison across all methods and all 16 PROCGEN environments (mean $\pm$ std, in \%). LAFP achieves the highest average performance, with particularly pronounced gains in environments requiring diverse or multimodal behaviors.}
  \label{tab:main-results}
  \centering
  \small
  \begin{tabular}{lcccc}
    \toprule
    \multirow{2}[1]{*}{\centering\textbf{Environment}} & \multicolumn{4}{c}{\textbf{Method}} \\
    \cmidrule(l){2-5}
    & \textbf{LAOM} & \textbf{LAOM (Frozen)} & \textbf{LAFP (Fine-tuned)} & \textbf{LAFP} \\
    \midrule
    bigfish    & 89.4 $\pm$ 1.8 & 88.0 $\pm$ 3.9 & 82.4 $\pm$ 1.8 & 88.8 $\pm$ 2.9 \\
    bossfight  & 45.8 $\pm$ 2.8 & 50.4 $\pm$ 4.2 & 34.2 $\pm$ 4.3 & 55.0 $\pm$ 7.1 \\
    caveflyer  & 16.8 $\pm$ 2.6 & 19.4 $\pm$ 4.4 & 2.2  $\pm$ 2.3 & 27.8 $\pm$ 3.6 \\
    chaser     & 19.8 $\pm$ 3.4 & 8.8  $\pm$ 2.8 & 27.0 $\pm$ 1.4 & 29.6 $\pm$ 7.4 \\
    climber    & 17.6 $\pm$ 4.2 & 20.4 $\pm$ 4.6 & 24.0 $\pm$ 4.2 & 29.0 $\pm$ 5.9 \\
    coinrun    & 93.4 $\pm$ 1.9 & 94.0 $\pm$ 2.7 & 94.4 $\pm$ 1.8 & 96.4 $\pm$ 1.1 \\
    dodgeball  & 35.0 $\pm$ 4.7 & 29.6 $\pm$ 4.3 & 0.0  $\pm$ 0.0 & 28.2 $\pm$ 6.2 \\
    fruitbot   & 39.2 $\pm$ 3.7 & 42.6 $\pm$ 6.0 & 33.0 $\pm$ 5.1 & 35.6 $\pm$ 3.2 \\
    heist      & 83.4 $\pm$ 4.0 & 83.2 $\pm$ 1.5 & 93.4 $\pm$ 0.9 & 97.2 $\pm$ 1.1 \\
    jumper     & 66.4 $\pm$ 3.2 & 69.6 $\pm$ 3.0 & 81.6 $\pm$ 4.5 & 79.4 $\pm$ 7.2 \\
    leaper     & 70.4 $\pm$ 5.5 & 71.2 $\pm$ 2.6 & 62.2 $\pm$ 1.6 & 73.0 $\pm$ 2.2 \\
    maze       & 98.4 $\pm$ 1.3 & 98.0 $\pm$ 1.7 & 93.0 $\pm$ 2.0 & 99.8 $\pm$ 0.4 \\
    miner      & 36.4 $\pm$ 4.2 & 31.8 $\pm$ 5.2 & 84.8 $\pm$ 1.3 & 87.0 $\pm$ 2.5 \\
    ninja      & 75.0 $\pm$ 3.4 & 73.2 $\pm$ 3.7 & 88.4 $\pm$ 2.4 & 92.4 $\pm$ 3.0 \\
    plunder    & 8.6  $\pm$ 2.9 & 5.8  $\pm$ 1.6 & 9.4  $\pm$ 4.6 & 8.0  $\pm$ 2.1 \\
    starpilot  & 75.0 $\pm$ 2.1 & 76.2 $\pm$ 3.8 & 0.0  $\pm$ 0.0 & 73.6 $\pm$ 4.6 \\
    \midrule
    Average    & 54.4 & 53.9 & 50.6 & 62.6 \\
    \bottomrule
  \end{tabular}
\end{table}

\subsection{Quantitative Latent Space Alignment}
\label{appendix:mmd}

Table~\ref{tab:mmd-results} reports squared MMD between each frozen policy's latent action distribution and the pretrained IDM reference on the same held-out observations. LAFP is closer to the IDM than LAOM(Frozen) in every environment, with an average of 0.00060 versus 0.03846. This full dimensional measure complements the UMAP visualizations reported in the main paper.

\begin{table}[htbp]
  \caption{Squared MMD between each distilled policy latent distribution and the pretrained IDM reference across all 16 PROCGEN environments. Lower values indicate better structure preservation. Values are averaged over five random seeds.}
  \label{tab:mmd-results}
  \centering
  \small
  \begin{tabular}{lcc@{\qquad}lcc}
    \toprule
    Environment & LAFP & LAOM(Frozen) & Environment & LAFP & LAOM(Frozen) \\
    \midrule
    BigFish   & 0.00055 & 0.03262 & Heist     & 0.00021 & 0.05584 \\
    BossFight & 0.00048 & 0.01464 & Jumper    & 0.00061 & 0.06160 \\
    CaveFlyer & 0.00045 & 0.04481 & Leaper    & 0.00035 & 0.03780 \\
    Chaser    & 0.00051 & 0.00692 & Maze      & 0.00025 & 0.00375 \\
    Climber   & 0.00092 & 0.01648 & Miner     & 0.00070 & 0.08450 \\
    CoinRun   & 0.00110 & 0.00530 & Ninja     & 0.00047 & 0.01473 \\
    Dodgeball & 0.00021 & 0.01428 & Plunder   & 0.00099 & 0.16817 \\
    FruitBot  & 0.00130 & 0.04756 & StarPilot & 0.00068 & 0.00679 \\
    \midrule
    \multicolumn{6}{c}{
      \textbf{Average}\quad
      LAFP: \textbf{0.00060}
      \qquad
      LAOM(Frozen): \textbf{0.03846}
    } \\
    \bottomrule
  \end{tabular}
\end{table}

\subsection{Overall Performance}
Table~\ref{tab:main-results} summarizes the performance of all methods. Overall, the results consistently corroborate the conclusions drawn in the main paper. LAFP achieves superior performance over LAOM in the majority of environments, with particularly notable gains in tasks that require diverse or multimodal behaviors, such as \textit{chaser}, \textit{jumper} and \textit{miner}. In contrast, in relatively simple environments such as \textit{coinrun} and \textit{maze}, both methods perform comparably, suggesting that the advantage of LAFP mainly arises in scenarios where preserving action diversity is critical.

Across all environments, LAFP benefits from freezing the distilled latent policy during action decoder post-training. LAFP(Fine-tuned) updates the flow policy jointly with the Post-AD and decreases from 62.6\% to 50.6\% on average, including 0\% success in Dodgeball and StarPilot. Both tasks are sensitive to individual action errors and offer few alternative valid actions, leaving limited room for recovery once fine-tuning perturbs the pretrained latent geometry. By contrast, fine-tuning changes LAOM only slightly because its pointwise mapping can adapt directly to the labeled action objective.

\subsection{Additional Ablation Results}
\label{appendix:ablation_full}

\begin{table}[t!]
  \caption{Ablation on the Stage-1 IDM-AD constraint. AD denotes models trained with the constraint and NAD denotes models trained without it.}
  \label{tab:ad-results}
  \centering
  \resizebox{\textwidth}{!}{%
  \small
  \begin{tabular}{lcccccccc}
    \toprule
    \multirow{4}{*}{\textbf{Environment}} & \multicolumn{8}{c}{\textbf{Method}} \\
    \cmidrule(l){2-9}
    & \multicolumn{2}{c}{\textbf{LAOM}} & \multicolumn{2}{c}{\textbf{LAOM (Frozen)}} & \multicolumn{2}{c}{\textbf{LAFP (Fine-tuned)}} & \multicolumn{2}{c}{\textbf{LAFP}} \\
    \cmidrule(lr){2-3} \cmidrule(lr){4-5} \cmidrule(lr){6-7} \cmidrule(l){8-9}
    & AD & NAD & AD & NAD & AD & NAD & AD & NAD \\
    \midrule
    bigfish    & 89.4 $\pm$ 1.8 & 85.2 $\pm$ 6.1 & 88.0 $\pm$ 3.9 & 88.0 $\pm$ 3.0 & 82.4 $\pm$ 1.8 & 65.6 $\pm$ 5.5 & 88.8 $\pm$ 2.9 & 79.6 $\pm$ 6.1 \\
    bossfight  & 45.8 $\pm$ 2.8 & 13.8 $\pm$ 1.3 & 50.4 $\pm$ 4.2 & 3.0  $\pm$ 0.7 & 34.2 $\pm$ 4.3 & 0.8  $\pm$ 0.8 & 55.0 $\pm$ 7.1 & 0.8  $\pm$ 1.3 \\
    caveflyer  & 16.8 $\pm$ 2.6 & 13.2 $\pm$ 4.1 & 19.4 $\pm$ 4.4 & 13.2 $\pm$ 4.0 & 2.2  $\pm$ 2.3 & 9.8  $\pm$ 3.6 & 27.8 $\pm$ 3.6 & 17.2 $\pm$ 4.1 \\
    chaser     & 19.8 $\pm$ 3.4 & 25.8 $\pm$ 3.5 & 8.8  $\pm$ 2.8 & 4.2  $\pm$ 1.6 & 27.0 $\pm$ 1.4 & 0.8  $\pm$ 0.8 & 29.6 $\pm$ 7.4 & 28.6 $\pm$ 3.5 \\
    climber    & 17.6 $\pm$ 4.2 & 22.8 $\pm$ 3.3 & 20.4 $\pm$ 4.6 & 20.4 $\pm$ 4.6 & 24.0 $\pm$ 4.2 & 24.6 $\pm$ 3.2 & 29.0 $\pm$ 5.9 & 21.2 $\pm$ 3.3 \\
    coinrun    & 93.4 $\pm$ 1.9 & 88.0 $\pm$ 2.6 & 94.0 $\pm$ 2.7 & 75.0 $\pm$ 3.4 & 94.4 $\pm$ 1.8 & 48.2 $\pm$ 4.7 & 96.4 $\pm$ 1.1 & 87.8 $\pm$ 2.6 \\
    dodgeball  & 35.0 $\pm$ 4.7 & 31.6 $\pm$ 3.6 & 29.6 $\pm$ 4.3 & 37.0 $\pm$ 1.6 & 0.0  $\pm$ 0.0 & 28.4 $\pm$ 5.8 & 28.2 $\pm$ 6.2 & 29.6 $\pm$ 3.6 \\
    fruitbot   & 39.2 $\pm$ 3.7 & 38.4 $\pm$ 4.7 & 42.6 $\pm$ 6.0 & 34.0 $\pm$ 4.3 & 33.0 $\pm$ 5.1 & 18.8 $\pm$ 5.5 & 35.6 $\pm$ 3.2 & 31.2 $\pm$ 4.7 \\
    heist      & 83.4 $\pm$ 4.0 & 83.0 $\pm$ 4.9 & 83.2 $\pm$ 1.5 & 50.0 $\pm$ 7.8 & 93.4 $\pm$ 0.9 & 88.6 $\pm$ 3.8 & 97.2 $\pm$ 1.1 & 69.6 $\pm$ 4.9 \\
    jumper     & 66.4 $\pm$ 3.2 & 71.4 $\pm$ 3.0 & 69.6 $\pm$ 3.0 & 67.6 $\pm$ 5.2 & 81.6 $\pm$ 4.5 & 73.8 $\pm$ 3.1 & 79.4 $\pm$ 7.2 & 77.6 $\pm$ 3.0 \\
    leaper     & 70.4 $\pm$ 5.5 & 74.8 $\pm$ 3.5 & 71.2 $\pm$ 2.6 & 72.2 $\pm$ 3.3 & 62.2 $\pm$ 1.6 & 71.8 $\pm$ 4.1 & 73.0 $\pm$ 2.2 & 72.0 $\pm$ 3.5 \\
    maze       & 98.4 $\pm$ 1.3 & 98.2 $\pm$ 1.6 & 98.0 $\pm$ 1.7 & 97.8 $\pm$ 1.3 & 93.0 $\pm$ 2.0 & 97.8 $\pm$ 0.8 & 99.8 $\pm$ 0.4 & 98.0 $\pm$ 1.6 \\
    miner      & 36.4 $\pm$ 4.2 & 48.2 $\pm$ 4.5 & 31.8 $\pm$ 5.2 & 41.0 $\pm$ 6.3 & 84.8 $\pm$ 1.3 & 78.6 $\pm$ 4.0 & 87.0 $\pm$ 2.5 & 72.8 $\pm$ 4.5 \\
    ninja      & 75.0 $\pm$ 3.4 & 79.8 $\pm$ 1.7 & 73.2 $\pm$ 3.7 & 77.4 $\pm$ 3.6 & 88.4 $\pm$ 2.4 & 78.0 $\pm$ 3.6 & 92.4 $\pm$ 3.0 & 81.0 $\pm$ 1.7 \\
    plunder    & 8.6  $\pm$ 2.9 & 3.0  $\pm$ 0.5 & 5.8  $\pm$ 1.6 & 1.6  $\pm$ 1.5 & 9.4  $\pm$ 4.6 & 0.0  $\pm$ 0.0 & 8.0  $\pm$ 2.1 & 0.4  $\pm$ 0.5 \\
    starpilot  & 75.0 $\pm$ 2.1 & 74.6 $\pm$ 3.6 & 76.2 $\pm$ 3.8 & 77.2 $\pm$ 3.8 & 0.0  $\pm$ 0.0 & 0.2  $\pm$ 0.4 & 73.6 $\pm$ 4.6 & 68.8 $\pm$ 3.6 \\
    \midrule
    Average    & 54.4 & 53.2 & 53.9 & 47.5 & 50.6 & 42.9 & 62.6 & 52.3 \\
    \bottomrule
  \end{tabular}%
  }
\end{table}

\begin{table}[t!]
  \caption{Comparison of decoder designs. Reusing the pretrained IDM-AD benefits LAFP but often degrades LAOM, highlighting the importance of preserving latent structure for flow-based policies.}
  \label{tab:idmad-results}
  \centering
  \small
  \begin{tabular}{lcccc}
    \toprule
    \multirow{4}{*}{\textbf{Environment}} & \multicolumn{4}{c}{\textbf{Method}} \\
    \cmidrule(l){2-5}
    & \multicolumn{2}{c}{\textbf{LAOM (Frozen)}} & \multicolumn{2}{c}{\textbf{LAFP}} \\
    \cmidrule(lr){2-3} \cmidrule(l){4-5}
    & Post-AD & IDM-AD & Post-AD & IDM-AD \\
    \midrule
    bigfish    & 88.0 $\pm$ 3.9 & 89.4 $\pm$ 3.4 & 88.8 $\pm$ 2.9 & 87.6 $\pm$ 3.2 \\
    bossfight  & 50.4 $\pm$ 4.2 & 47.0 $\pm$ 6.1 & 55.0 $\pm$ 7.1 & 46.0 $\pm$ 3.8 \\
    caveflyer  & 19.4 $\pm$ 4.4 & 19.4 $\pm$ 5.3 & 27.8 $\pm$ 3.6 & 33.0 $\pm$ 7.2 \\
    chaser     & 8.8  $\pm$ 2.8 & 5.6  $\pm$ 4.8 & 29.6 $\pm$ 7.4 & 25.6 $\pm$ 1.8 \\
    climber    & 20.4 $\pm$ 4.6 & 14.8 $\pm$ 3.2 & 29.0 $\pm$ 5.9 & 29.6 $\pm$ 0.8 \\
    coinrun    & 94.0 $\pm$ 2.7 & 97.8 $\pm$ 2.3 & 96.4 $\pm$ 1.1 & 97.8 $\pm$ 1.1 \\
    dodgeball  & 29.6 $\pm$ 4.3 & 20.2 $\pm$ 2.6 & 28.2 $\pm$ 6.2 & 33.4 $\pm$ 2.9 \\
    fruitbot   & 42.6 $\pm$ 6.0 & 38.8 $\pm$ 7.3 & 35.6 $\pm$ 3.2 & 36.6 $\pm$ 5.2 \\
    heist      & 83.2 $\pm$ 1.5 & 12.2 $\pm$ 1.5 & 97.2 $\pm$ 1.1 & 96.8 $\pm$ 3.0 \\
    jumper     & 69.6 $\pm$ 3.0 & 62.0 $\pm$ 6.3 & 79.4 $\pm$ 7.2 & 83.4 $\pm$ 5.8 \\
    leaper     & 71.2 $\pm$ 2.6 & 70.2 $\pm$ 3.7 & 73.0 $\pm$ 2.2 & 75.0 $\pm$ 3.7 \\
    maze       & 98.0 $\pm$ 1.7 & 96.8 $\pm$ 0.8 & 99.8 $\pm$ 0.4 & 99.2 $\pm$ 1.9 \\
    miner      & 31.8 $\pm$ 5.2 & 0.2  $\pm$ 3.9 & 87.0 $\pm$ 2.5 & 81.4 $\pm$ 0.4 \\
    ninja      & 73.2 $\pm$ 3.7 & 63.8 $\pm$ 2.1 & 92.4 $\pm$ 3.0 & 93.0 $\pm$ 6.7 \\
    plunder    & 5.8  $\pm$ 1.6 & 4.4  $\pm$ 1.6 & 8.0  $\pm$ 2.1 & 6.2  $\pm$ 1.5 \\
    starpilot  & 76.2 $\pm$ 3.8 & 70.6 $\pm$ 2.7 & 73.6 $\pm$ 4.6 & 76.6 $\pm$ 0.9 \\
    \midrule
    Average    & 53.9 & 44.6 & 62.6 & 62.6 \\
    \bottomrule
  \end{tabular}
\end{table}

\begin{table}[t!]
  \caption{Effect of the number of flow steps. A small number of steps (3–5) achieves the best balance between performance and efficiency, while more steps yield diminishing returns.}
  \label{tab:steps-results}
  \centering
  \small
  \begin{tabular}{lccccc}
    \toprule
    \multirow{2}[1]{*}{\textbf{Environment}} & \multicolumn{5}{c}{\textbf{Model}} \\
    \cmidrule(l){2-6}
    & \textbf{LAFP(1-step)} & \textbf{LAFP(3-steps)} & \textbf{LAFP(5-steps)} & \textbf{LAFP(20-steps)} & \textbf{LAOM} \\
    \midrule
    bigfish    & 87.6 $\pm$ 4.5 & 88.8 $\pm$ 2.9 & 84.2 $\pm$ 4.8 & 84.8 $\pm$ 4.0 & 86.2 $\pm$ 1.9 \\
    bossfight  & 48.6 $\pm$ 2.4 & 55.0 $\pm$ 7.1 & 48.2 $\pm$ 3.3 & 51.4 $\pm$ 6.3 & 46.6 $\pm$ 2.1 \\
    caveflyer  & 20.0 $\pm$ 5.3 & 27.8 $\pm$ 3.6 & 28.8 $\pm$ 4.1 & 27.0 $\pm$ 4.5 & 17.4 $\pm$ 4.3 \\
    chaser     & 29.0 $\pm$ 1.9 & 29.6 $\pm$ 7.4 & 23.2 $\pm$ 5.8 & 15.2 $\pm$ 3.8 & 17.0 $\pm$ 2.1 \\
    climber    & 21.6 $\pm$ 3.8 & 29.0 $\pm$ 5.9 & 27.8 $\pm$ 2.0 & 28.8 $\pm$ 4.0 & 15.4 $\pm$ 2.2 \\
    coinrun    & 97.4 $\pm$ 1.8 & 96.4 $\pm$ 1.1 & 95.2 $\pm$ 2.5 & 96.6 $\pm$ 2.5 & 92.6 $\pm$ 3.2 \\
    dodgeball  & 32.4 $\pm$ 7.1 & 28.2 $\pm$ 6.2 & 26.8 $\pm$ 2.6 & 31.8 $\pm$ 2.9 & 31.2 $\pm$ 1.9 \\
    fruitbot   & 41.8 $\pm$ 7.0 & 35.6 $\pm$ 3.2 & 35.0 $\pm$ 5.3 & 33.6 $\pm$ 3.9 & 40.8 $\pm$ 6.4 \\
    heist      & 88.6 $\pm$ 2.5 & 97.2 $\pm$ 1.1 & 97.6 $\pm$ 0.9 & 95.2 $\pm$ 1.6 & 81.4 $\pm$ 2.7 \\
    jumper     & 71.4 $\pm$ 3.8 & 79.4 $\pm$ 7.2 & 84.8 $\pm$ 2.9 & 82.4 $\pm$ 4.3 & 65.4 $\pm$ 3.2 \\
    leaper     & 71.0 $\pm$ 4.0 & 73.0 $\pm$ 2.2 & 72.6 $\pm$ 2.4 & 73.4 $\pm$ 6.2 & 68.0 $\pm$ 4.3 \\
    maze       & 98.4 $\pm$ 0.9 & 99.8 $\pm$ 0.4 & 99.8 $\pm$ 0.4 & 99.6 $\pm$ 0.5 & 98.0 $\pm$ 0.7 \\
    miner      & 57.4 $\pm$ 4.7 & 87.0 $\pm$ 2.5 & 86.6 $\pm$ 2.3 & 86.6 $\pm$ 5.7 & 34.6 $\pm$ 5.4 \\
    ninja      & 70.6 $\pm$ 5.9 & 92.4 $\pm$ 3.0 & 92.2 $\pm$ 2.6 & 92.2 $\pm$ 2.0 & 75.0 $\pm$ 2.1 \\
    plunder    & 7.6  $\pm$ 2.1 & 8.0  $\pm$ 2.1 & 7.0  $\pm$ 2.4 & 5.6  $\pm$ 3.0 & 9.0  $\pm$ 1.6 \\
    starpilot  & 74.8 $\pm$ 4.7 & 73.6 $\pm$ 4.6 & 73.0 $\pm$ 3.5 & 71.2 $\pm$ 3.1 & 77.2 $\pm$ 4.2 \\
    \midrule
    Average    & 57.4 & 62.6 & 61.4 & 61.0 & 53.5 \\
    \midrule
    \textbf{\makecell[l]{Inference time \\ per action (ms)}} & 1.59 & 2.00 & 2.38 & 5.51 & 1.09 \\
    \bottomrule
  \end{tabular}
\end{table}

\begin{table}[t!]
  \caption{Effect of latent dimension and prediction target. Direct latent prediction remains robust across dimensions, while vector field prediction degrades significantly in high dimensional settings.}
  \label{tab:xv-results}
  \centering
  \resizebox{\textwidth}{!}{%
  \begin{tabular}{lcccccccc}
    \toprule
    \multirow{4}{*}{\textbf{Environment}} & \multicolumn{8}{c}{\textbf{Dimension \& Predicted Target}} \\
    \cmidrule(l){2-9}
    & \multicolumn{2}{c}{\textbf{32}} & \multicolumn{2}{c}{\textbf{64}} & \multicolumn{2}{c}{\textbf{128}} & \multicolumn{2}{c}{\textbf{256}} \\
    \cmidrule(lr){2-3} \cmidrule(lr){4-5} \cmidrule(lr){6-7} \cmidrule(l){8-9}
    & latent & vector & latent & vector & latent & vector & latent & vector \\
    \midrule
    bigfish    & 85.4 $\pm$ 5.3 & 88.4 $\pm$ 2.5 & 87.4 $\pm$ 3.2 & 87.4 $\pm$ 4.0 & 88.8 $\pm$ 2.9 & 86.4 $\pm$ 4.1 & 86.0 $\pm$ 2.9 & 38.4 $\pm$ 3.6 \\
    bossfight  & 55.0 $\pm$ 8.1 & 53.4 $\pm$ 4.3 & 48.4 $\pm$ 4.0 & 52.0 $\pm$ 3.2 & 55.0 $\pm$ 7.1 & 49.4 $\pm$ 4.5 & 46.6 $\pm$ 3.3 & 49.6 $\pm$ 4.4 \\
    caveflyer  & 28.6 $\pm$ 5.2 & 30.2 $\pm$ 4.7 & 28.8 $\pm$ 2.9 & 27.0 $\pm$ 2.7 & 27.8 $\pm$ 3.6 & 28.8 $\pm$ 5.4 & 29.4 $\pm$ 2.1 & 21.8 $\pm$ 4.8 \\
    chaser     & 35.2 $\pm$ 4.7 & 31.6 $\pm$ 5.2 & 27.0 $\pm$ 3.8 & 26.2 $\pm$ 3.7 & 29.6 $\pm$ 7.4 & 28.8 $\pm$ 3.1 & 26.8 $\pm$ 3.8 & 0.0  $\pm$ 0.0 \\
    climber    & 25.6 $\pm$ 5.3 & 23.0 $\pm$ 3.1 & 26.4 $\pm$ 6.0 & 23.6 $\pm$ 6.5 & 29.0 $\pm$ 5.9 & 25.0 $\pm$ 7.4 & 27.4 $\pm$ 1.1 & 28.4 $\pm$ 5.0 \\
    coinrun    & 97.4 $\pm$ 1.1 & 94.2 $\pm$ 2.8 & 95.0 $\pm$ 1.4 & 93.0 $\pm$ 2.2 & 96.4 $\pm$ 1.1 & 97.0 $\pm$ 1.2 & 96.8 $\pm$ 2.8 & 90.8 $\pm$ 4.1 \\
    dodgeball  & 29.6 $\pm$ 2.7 & 32.2 $\pm$ 6.5 & 30.8 $\pm$ 5.4 & 28.2 $\pm$ 1.9 & 28.2 $\pm$ 6.2 & 28.0 $\pm$ 5.6 & 29.0 $\pm$ 5.8 & 29.0 $\pm$ 5.8 \\
    fruitbot   & 37.0 $\pm$ 4.2 & 32.8 $\pm$ 4.8 & 37.2 $\pm$ 7.4 & 31.6 $\pm$ 3.4 & 35.6 $\pm$ 3.2 & 35.2 $\pm$ 4.1 & 39.8 $\pm$ 7.3 & 32.2 $\pm$ 2.2 \\
    heist      & 98.4 $\pm$ 1.5 & 98.8 $\pm$ 0.8 & 96.6 $\pm$ 1.5 & 96.4 $\pm$ 1.5 & 97.2 $\pm$ 1.1 & 94.2 $\pm$ 1.9 & 97.2 $\pm$ 1.9 & 95.4 $\pm$ 1.9 \\
    jumper     & 82.2 $\pm$ 1.3 & 81.2 $\pm$ 3.7 & 84.6 $\pm$ 0.9 & 82.4 $\pm$ 3.8 & 79.4 $\pm$ 7.2 & 79.2 $\pm$ 2.9 & 83.6 $\pm$ 5.4 & 78.4 $\pm$ 2.9 \\
    leaper     & 71.6 $\pm$ 3.9 & 72.8 $\pm$ 2.9 & 72.4 $\pm$ 3.5 & 73.8 $\pm$ 5.5 & 73.0 $\pm$ 2.2 & 71.4 $\pm$ 2.7 & 73.6 $\pm$ 4.4 & 71.6 $\pm$ 4.8 \\
    maze       & 99.4 $\pm$ 0.9 & 99.6 $\pm$ 0.5 & 98.4 $\pm$ 1.9 & 100.0 $\pm$ 0.0 & 99.8 $\pm$ 0.4 & 99.6 $\pm$ 0.5 & 99.6 $\pm$ 0.5 & 100.0 $\pm$ 0.0 \\
    miner      & 88.4 $\pm$ 2.1 & 83.6 $\pm$ 5.4 & 84.4 $\pm$ 1.9 & 83.4 $\pm$ 1.7 & 87.0 $\pm$ 2.5 & 77.0 $\pm$ 6.7 & 86.4 $\pm$ 4.8 & 15.8 $\pm$ 2.6 \\
    ninja      & 94.8 $\pm$ 1.8 & 90.6 $\pm$ 2.2 & 93.2 $\pm$ 2.5 & 90.2 $\pm$ 0.8 & 92.4 $\pm$ 3.0 & 92.8 $\pm$ 3.7 & 92.6 $\pm$ 3.0 & 90.2 $\pm$ 2.3 \\
    plunder    & 7.0  $\pm$ 2.0 & 1.8  $\pm$ 1.9 & 9.0  $\pm$ 4.2 & 3.6  $\pm$ 1.7 & 8.0  $\pm$ 2.1 & 2.2  $\pm$ 1.1 & 6.8  $\pm$ 2.4 & 0.2  $\pm$ 0.4 \\
    starpilot  & 75.8 $\pm$ 2.7 & 77.4 $\pm$ 6.1 & 74.6 $\pm$ 3.8 & 75.2 $\pm$ 2.0 & 73.6 $\pm$ 4.6 & 73.6 $\pm$ 3.2 & 77.4 $\pm$ 2.1 & 69.0 $\pm$ 5.5 \\
    \midrule
    \textbf{Average} & 63.2 & 62.0 & 62.1 & 60.9 & 62.6 & 60.5 & 62.4 & 50.7 \\
    \bottomrule
  \end{tabular}%
  }
\end{table}

We report all ablations over five random seeds and 100 episodes per seed; standard deviations are computed across seeds. Table~\ref{tab:ad-results} shows that the Stage-1 IDM-AD constraint is particularly important for LAFP. Table~\ref{tab:idmad-results} evaluates whether the frozen IDM-AD can be transferred after Stage~2, providing a functional test of latent space alignment. Table~\ref{tab:steps-results} shows that three to five flow steps provide the best performance--efficiency trade-off. Table~\ref{tab:xv-results} shows that x-prediction remains robust as the latent dimension grows, whereas v-prediction degrades in the highest dimensional setting.

\begin{figure}[htbp]
    \centering
    \fbox{\begin{minipage}{\dimexpr0.89\textwidth-2\fboxsep-2\fboxrule\relax}
        \centering
        \includegraphics[width=\textwidth]{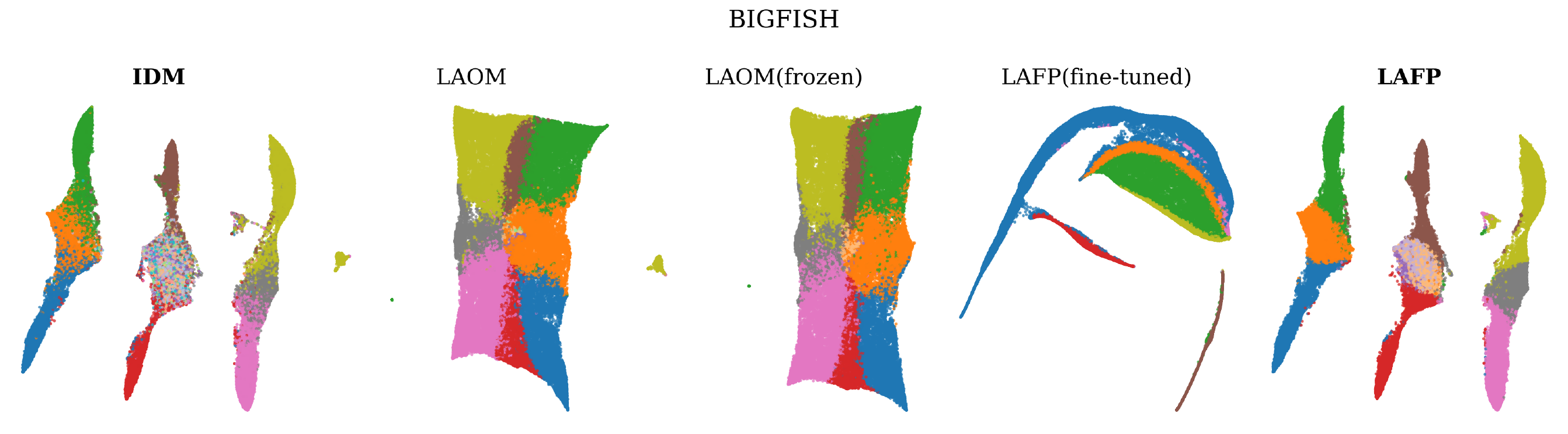}\\[2pt]
        \includegraphics[width=\textwidth]{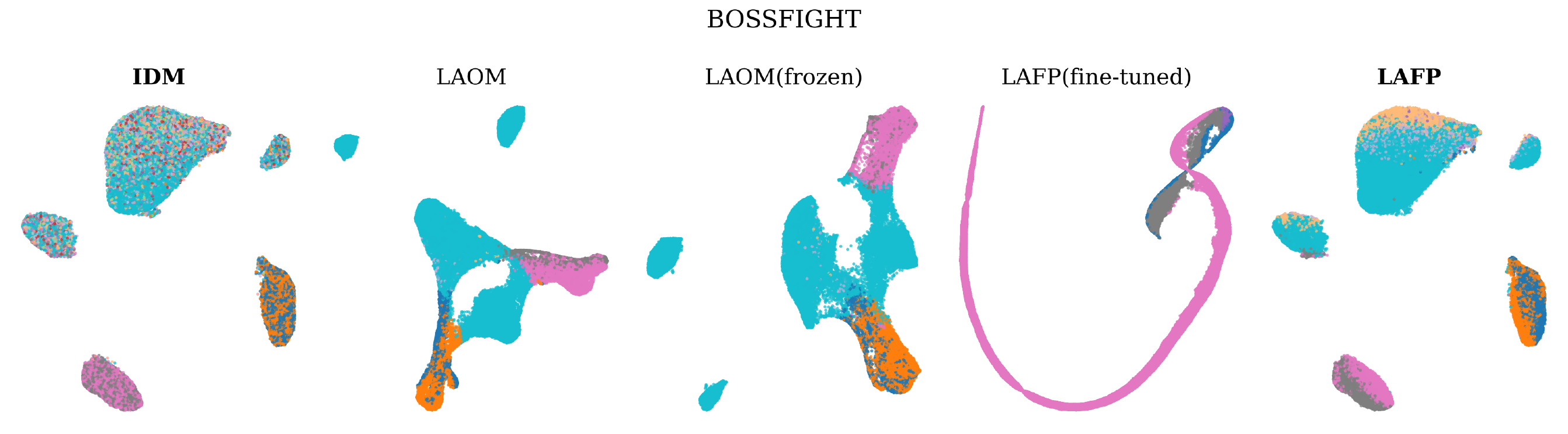}\\[2pt]
        \includegraphics[width=\textwidth]{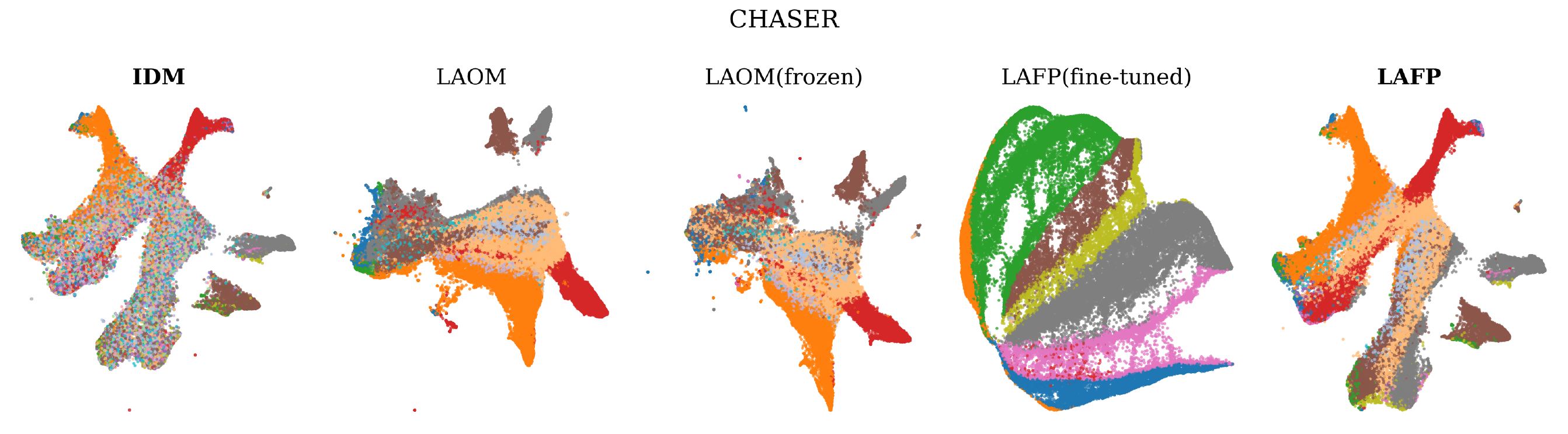}\\[2pt]
        \includegraphics[width=\textwidth]{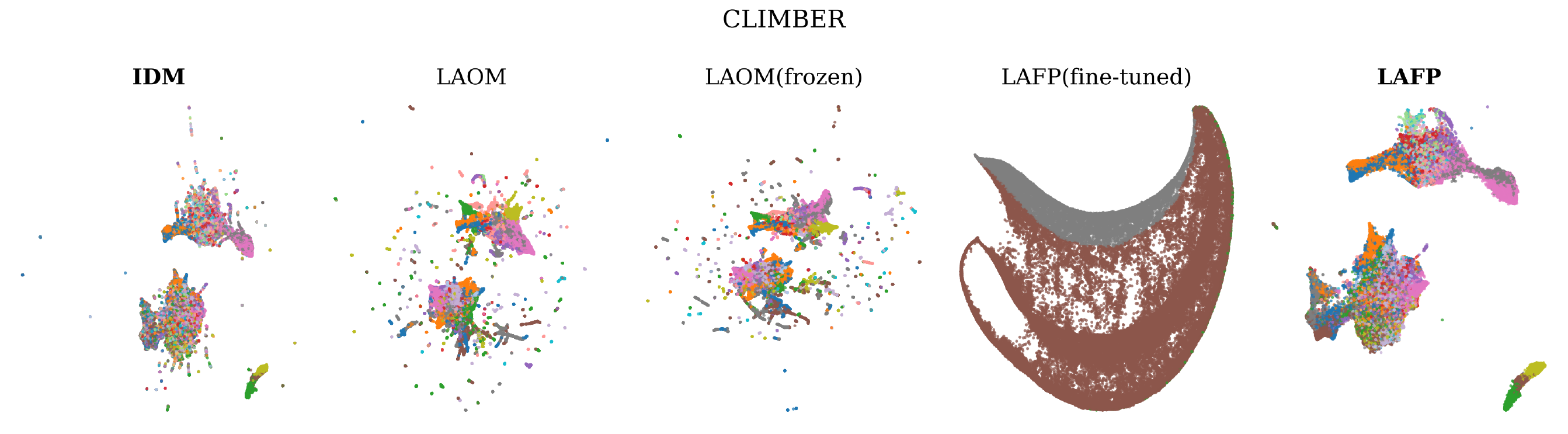}\\[2pt]
        \includegraphics[width=\textwidth]{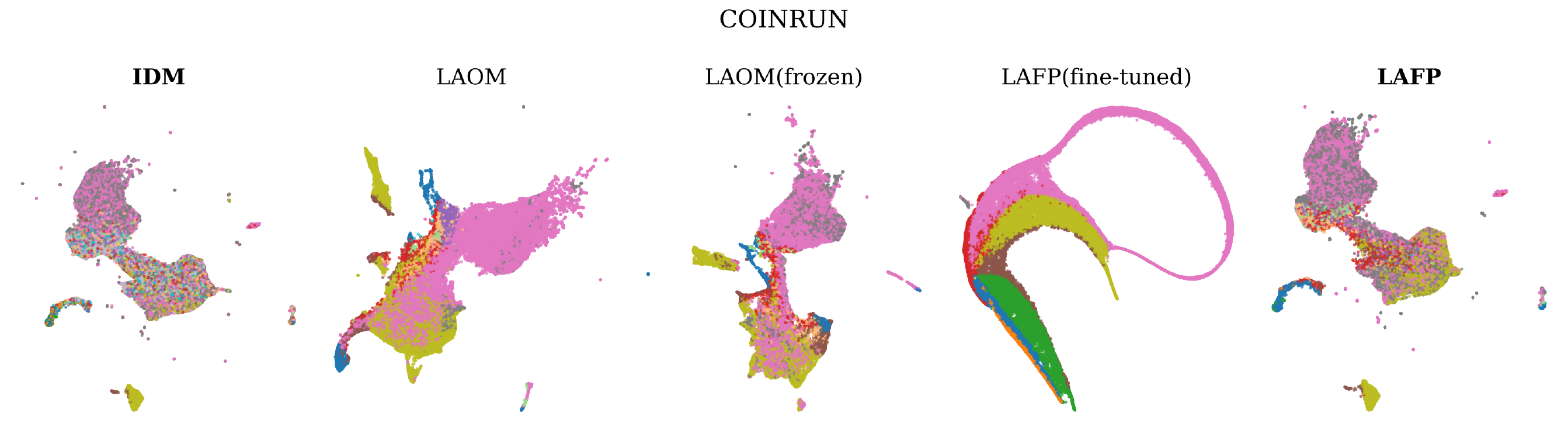}\\[2pt]
        \includegraphics[width=\textwidth]{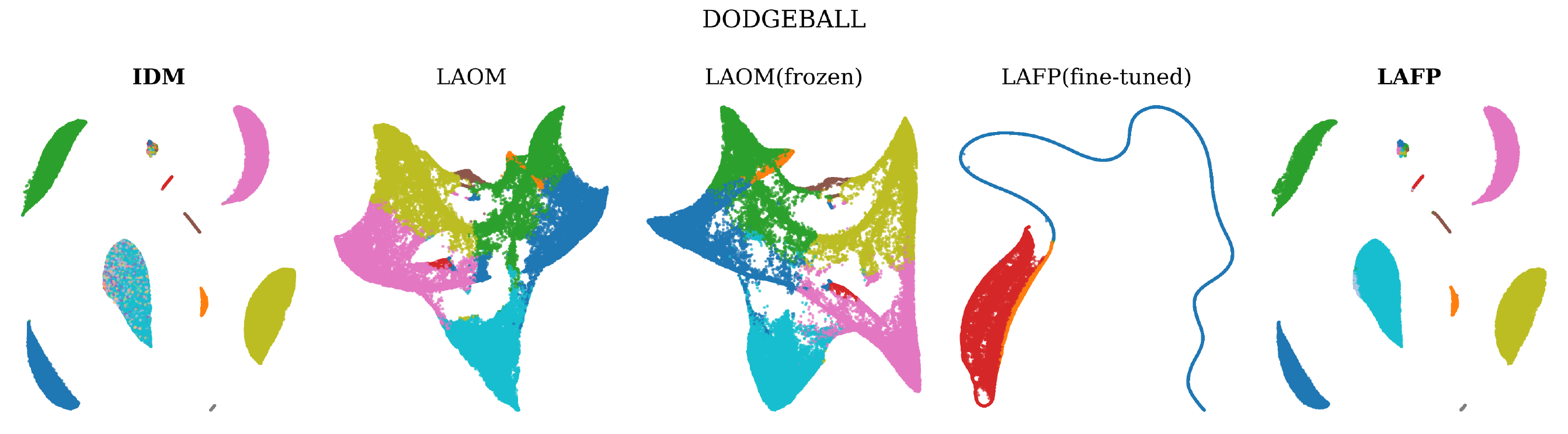}\\[2pt]
    \end{minipage}}
    \caption*{}
\end{figure}


\begin{figure}[htbp]
    \centering
    \fbox{\begin{minipage}{\dimexpr0.89\textwidth-2\fboxsep-2\fboxrule\relax}
        \centering
        \includegraphics[width=\textwidth]{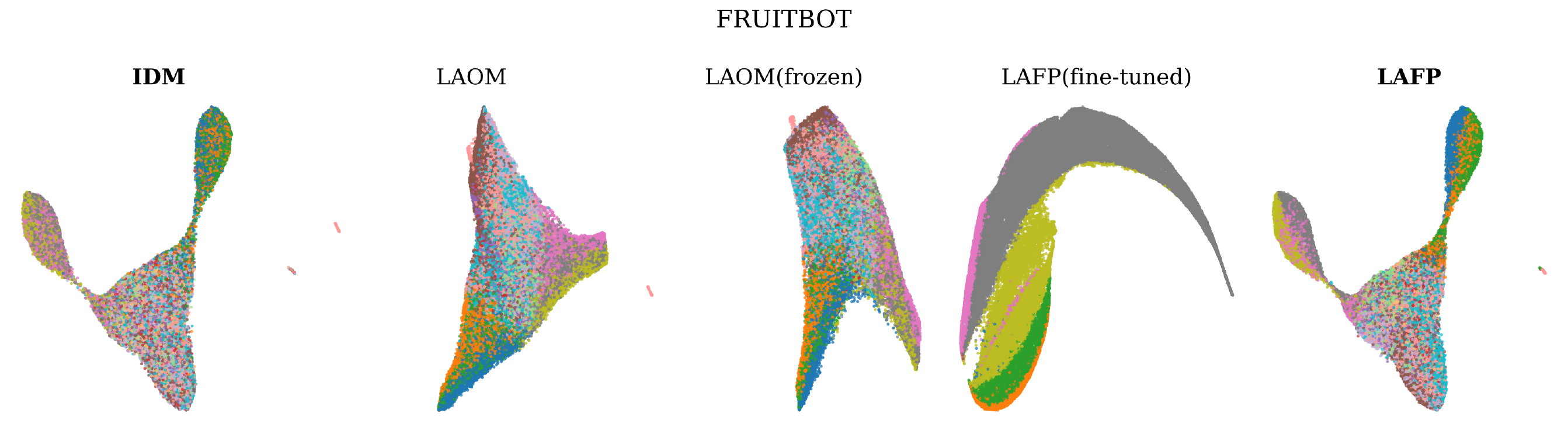}\\[2pt]
        \includegraphics[width=\textwidth]{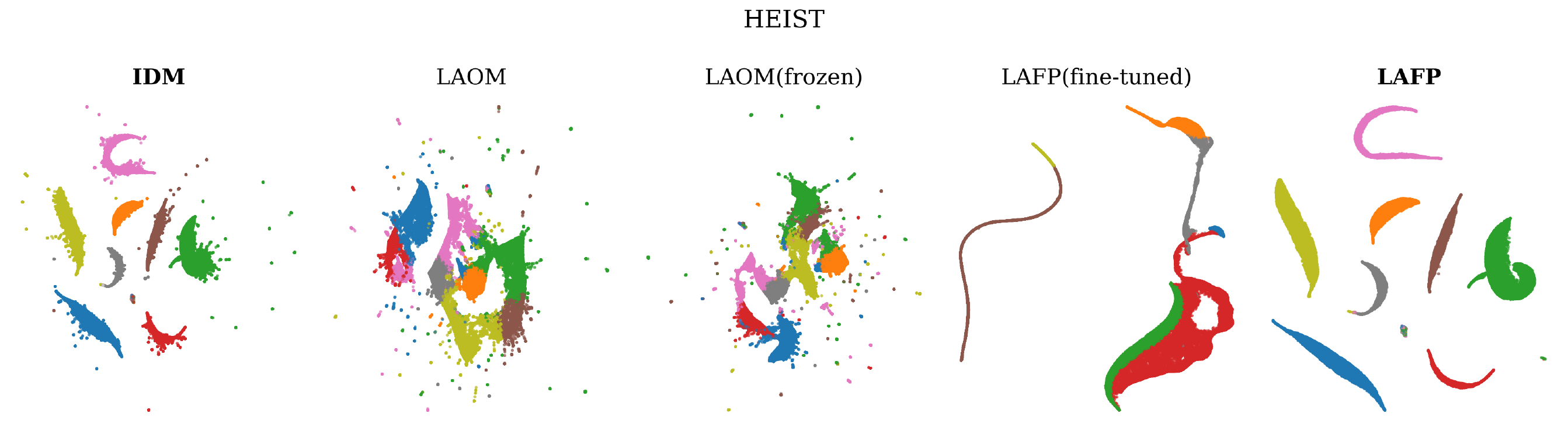}\\[2pt]
        \includegraphics[width=\textwidth]{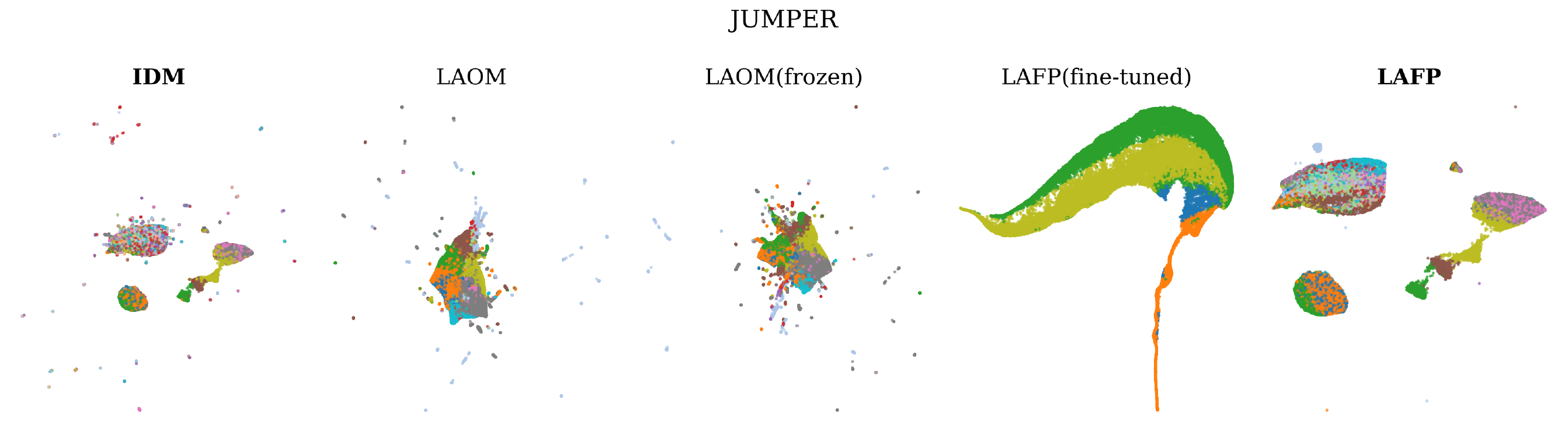}\\[2pt]
        \includegraphics[width=\textwidth]{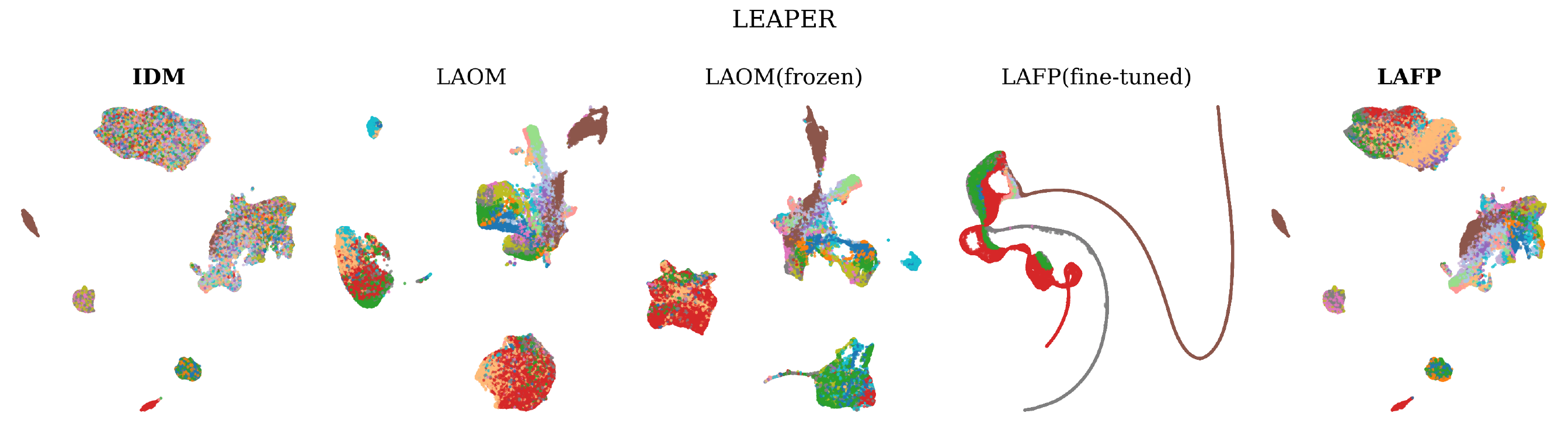}\\[2pt]
        \includegraphics[width=\textwidth]{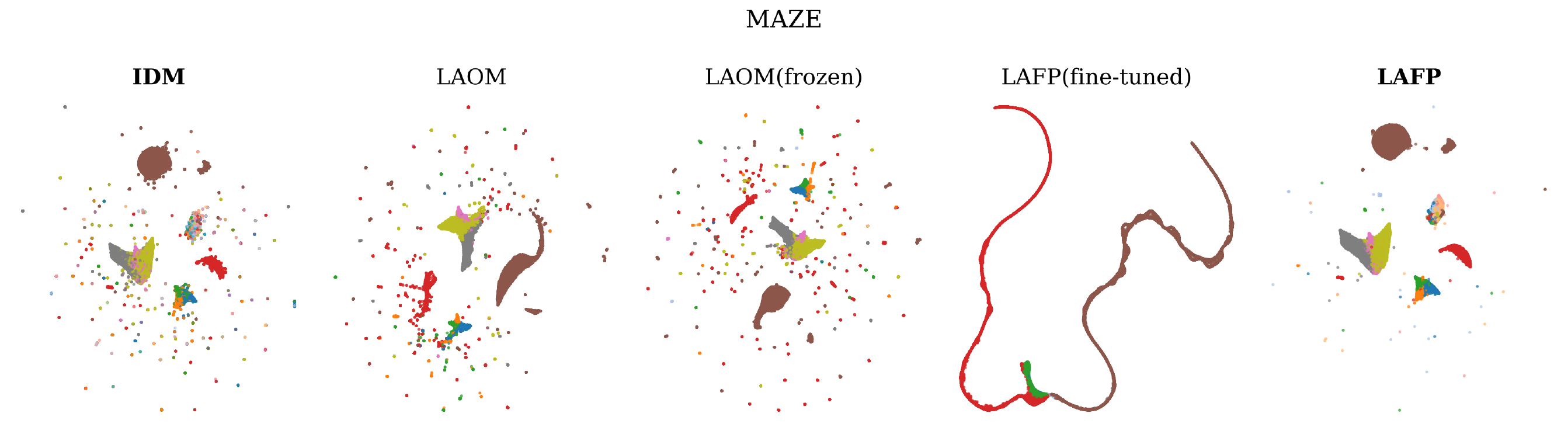}\\[2pt]
        \includegraphics[width=\textwidth]{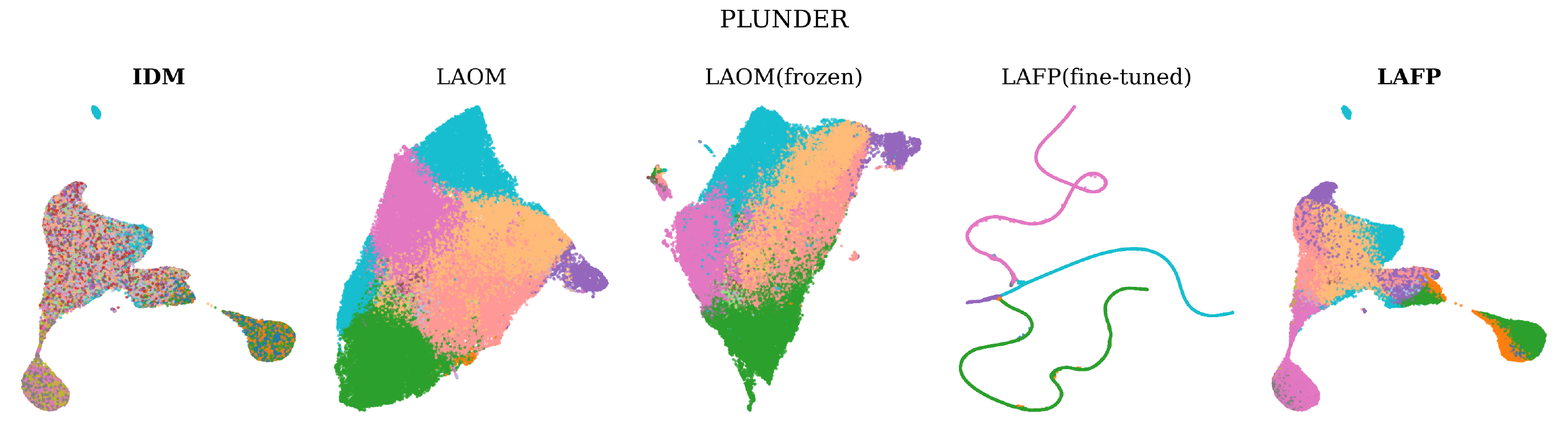}\\[2pt]
        \includegraphics[width=0.5\textwidth]{figures/umap_color_bar.png}
    \end{minipage}}
    \caption{UMAP projections of latent action spaces for the IDM and downstream policies across the remaining 12 PROCGEN environments.}  
    \label{fig:IDM bc flow umap all}
\end{figure}

\section{Additional UMAP Visualizations}
\label{appendix:umap_full}
Figure~\ref{fig:IDM bc flow umap all} provides UMAP visualizations for the remaining environments. The IDM forms clusters associated with different action modes. LAFP most closely retains this organization, whereas LAOM and LAFP(Fine-tuned) tend to distort or contract multiple modes. UMAP is only qualitative; the full dimensional MMD results in Table~\ref{tab:mmd-results} and the IDM-AD transfer test provide the corresponding quantitative evidence.

\section{Implementation Details}
\label{appendix:details}

\paragraph{Models.}
All methods share architectures and optimization settings across environments, without per-environment tuning. Our implementation builds on the official LAPO codebase. The IDM uses a lightweight CNN encoder, the FDM uses a U-Net, and the flow policy is a three-layer MLP conditioned on CNN observation features and a sinusoidal embedding of flow time $\tau$. Both IDM-AD and Post-AD are three-layer MLPs.

\paragraph{Normalization.}
After pre-training, we compute per-dimension statistics of the latent action space for each environment. Specifically, the center (midrange) is defined as the average of the maximum and minimum values along each dimension, and the standard deviation is computed accordingly. These statistics are used to normalize latent actions during both training and inference in subsequent stages, which improves numerical stability and ensures consistent scaling across environments.

\paragraph{Computational cost.}
All experiments are conducted on a single NVIDIA H20 GPU per environment under the same computational budget. In terms of computational cost, pre-training takes approximately 50 minutes per environment. Distillation requires around one hour for LAFP and 40-50 minutes for LAOM (behavior cloning), suggesting that the additional cost of flow matching is relatively minor. The post-training stage is lightweight, requiring only a few minutes. Overall, the total training time per environment remains within a practical range.

\end{document}